\documentclass[11pt]{article}

\usepackage[final]{acl}

\usepackage{times}
\usepackage{latexsym}

\usepackage[T1]{fontenc}

\usepackage[utf8]{inputenc}

\usepackage{microtype}

\usepackage{inconsolata}

\usepackage{graphicx}

\usepackage{booktabs}
\usepackage{listings}
\usepackage{amsmath}
\usepackage{multirow}
\usepackage{makecell}
\usepackage{siunitx}
\usepackage{threeparttable}
\usepackage{arydshln} 
\usepackage{enumitem}
\usepackage{tcolorbox}
 
\title{Automated Data Enrichment using Confidence-Aware Fine-Grained Debate among Open-Weight LLMs for Mental Health and Online Safety}

\author{
  \begin{tabular}[t]{c}
  \small\bfseries
  Junyu Mao$^{1}$ \quad
  Anthony Hills$^{2}$ \quad
  Talia Tseriotou$^{2}$ \quad
  Maria Liakata$^{2,3}$ \quad
  Aya Shamir$^{4}$ \quad
  Dan Sayda$^{4}$ \quad
  \\
  \small\bfseries
  Dana Atzil-Slonim$^{4}$ \quad
  Natalie Djohari$^{1}$ \quad
  Pamela Ugwudike$^{1}$ \quad
  \\
  \small\bfseries
  Mahesan Niranjan$^{1}$ \quad
  Stuart E. Middleton$^{1}$ \quad
  \end{tabular}
  \\[2ex]
  \small
  $^{1}$University of Southampton, UK \\
  \small
  $^{2}$Queen Mary University of London, UK $^{3}$The Alan Turing Institute, UK  \\
  \small
  $^{4}$Bar Ilan University, Israel \\
  \small
  \texttt{\{junyu.mao, sem03\}@soton.ac.uk},
  \texttt{mn@ecs.soton.ac.uk}
}

\begin{document}
\maketitle

\begin{abstract}
Real-world indicators play an important role in many Natural Language Processing (NLP) applications, such as life events for mental health analysis and risky behaviours for online safety, yet labelling such information is often costly and/or difficult due to its multi-label and dynamic nature. Large Language Models (LLMs) show promising potential for automated annotation, but the multi-label setting remains challenging. In this work, we propose a Confidence-Aware Fine-Grained Debate (CFD) framework that simulates human collaborative annotation using fine-grained communication to better support automated multi-label enrichment. We introduce two expert-annotated resources: life-event and symptom annotations for a mental health well-being dataset, and a new online safety sharenting dataset. Experiments show that CFD achieves the most robust enrichment performance across tasks, and that the benefit of fine-grained confidence is influenced by its quality and variability. We further evaluate training-free strategies for incorporating enrichment indicators into downstream tasks and show that automated enrichment consistently improves performance, by up to 9.9 Macro-F1 points, with the most effective integration format depending on how the indicator relates to the downstream objective.
\end{abstract}

\section{Introduction}
In recent years, NLP has advanced rapidly, particularly with the rise of Large Language Models (LLMs) \citep{Brown2020gpt3,llama}. The success of LLMs has driven growing interest in applications across diverse domains such as mental health \citep{66-lamichhane2023evaluation,nguyen2024leveraginglargelanguagemodels,yang2023mentalllama,song-etal-2024-combining} and online safety \citep{Diaz_Garcia_2025}.

Beyond the original input, many task-related real-world indicators often provide valuable supplementary information that can not only improve performance on target tasks \citep{lan-etal-2025-depression} but also enhance interpretability \citep{chen-etal-2024-mapping}. However, many real-world events and behaviours are inherently multi-label, making annotation costly and time-consuming for domain experts, especially when annotators are scarce or when the label space is dynamic. Although LLMs show promising annotation capabilities without task-specific fine-tuning \citep{Gilardi_2023,törnberg2023chatgpt4outperformsexpertscrowd,alizadeh2023opensourcellmoutperform}, multi-label annotation remains challenging, particularly for open-weight models \citep{fillies2025mapping,cory2025wordlevelannotationgdprtransparency}.

In this paper, we propose a novel confidence-aware fine-grained debate (CFD) framework for automated data enrichment, in which LLM-based agents simulate a collaborative annotation team, leveraging the complementary strengths of open-weight LLMs. The framework combines categorical chain-of-thought reasoning, per-category debate, and fine-grained confidence communication to better support multi-label annotation. We study enrichment in two domains: \textbf{a)} for mental health, we enrich the existing CLPsych 2025 shared task dataset on well-being prediction of individuals from their social media posts \citep{tseriotou-etal-2025-overview} with life-event and symptom labels; and \textbf{b)} for online safety, we introduce a new sharenting benchmark, where sharenting \citep{roth2025sharenting} refers to sharing children's personal information online, which may unintentionally expose them to digital harms such as identity-related crimes, harassment and cyberbullying; posts are annotated with both risky behaviours and an overall risk level. CFD yields the most robust annotation performance across tasks, and our analysis shows that fine-grained confidence distinguishes correct from incorrect decisions far better than response-level confidence, with more marginal advantage when scores are more uniform across labels. Beyond annotation performance, we systematically evaluate training-free strategies for incorporating enrichment indicators, either as predicted labels or as reasoning traces, into the two downstream tasks, well-being score and sharenting risk prediction. Automated enrichment improves both tasks, gaining up to 9.9 Macro-F1 points, though which form works better depends on how the indicator relates to the downstream target. Our main contributions are summarised as follows:

\begin{itemize}[leftmargin=*,itemsep=2pt,parsep=0pt,topsep=2pt]
\item \textbf{We propose CFD}, a confidence-aware fine-grained debate framework for multi-label annotation with open-weight LLMs. To our knowledge, it is the first to debate confidence at a fine-grained level, and we systematically compare confidence communication across granularities.
\item \textbf{We release two new expert-annotated resources}\footnote{\url{https://github.com/milliemaoo/CFD}} to support NLP research on mental health and online safety: life-event and symptom labels for 350 Reddit posts from the CLPsych 2025 shared task dataset, and, to the best of our knowledge, the first sharenting risk benchmark with expert-annotated risk and risky behaviour labels on 1,901 Facebook posts.
\item \textbf{We compare different formats for integrating enrichment indicators into downstream tasks} without additional model tuning, and show consistent gains. Our results demonstrate the practical value of automated enrichment and provide baseline results for future research.
\end{itemize}

\section{Related work}
\paragraph{Automatic Annotation}
LLMs enable automated text annotation without task-specific fine-tuning, offering cost-effective and flexible alternatives when labelled data are scarce and manual labelling is expensive. In some cases, LLMs even outperform crowd workers \citep{Gilardi_2023,törnberg2023chatgpt4outperformsexpertscrowd,alizadeh2023opensourcellmoutperform}, though their performance can still be task-dependent \citep{ziems-etal-2024-llm-ComputationalSocialScience,ahmed2025llmsreplacemanualannotationSoftwareEngineeringArtifacts}. Compared with single-label annotation, multi-label annotation with LLMs remains relatively underexplored despite increasing interest across domains. \citet{hassan2024automatedmultilabelannotationmental} show that repeated
single-label prompting, applying the same prompt once per label, achieves
strong performance for mental illness annotation, though at high
computational cost. Other work adopts direct multi-label prompting, either through further fine-tuning \citep{fillies2025mapping} or multi-stage prompting with retrieval-augmented generation relying on a pool of illustrative examples \citep{cory2025wordlevelannotationgdprtransparency}. Despite these advances, direct multi-label annotation remains challenging, particularly for open-weight models. Our work focuses on improving this capability of open-weight LLMs under limited supervision by leveraging multi-LLM collaboration and incorporating fine-grained signals.

\paragraph{Multi-Agent Debate} 
Prior work explores multi-round interaction to enhance factuality and reasoning~\citep{du2023improvingfactualityreasoninglanguage}, role-playing communicative agents and diverse communication strategies~\citep{li2023camelcommunicativeagentsmind,chan2023chatevalbetterllmbasedevaluators}, as well as coordination mechanisms that foster divergent thinking~\citep{liang2024encouragingdivergentthinkinglarge}. Related setups also improve non-expert decision making by selecting answers from arguments proposed by expert LLMs~\citep{khan2024debatingpersuasivellmsleads}. Some work incorporates confidence into inter-agent communication, either through self-reported confidence via prompts \citep{chen2024reconcileroundtableconferenceimproves} or through uncertainty metrics that estimate confidence, conveyed via prompting or attention scaling \citep{yoffe2025debuncimprovinglargelanguage}. We also incorporate confidence into debate, but at a fine-grained level rather than a single response level.

\paragraph{Mental Health}
Recent research has explored language models for mental health applications on social media data \citep{hills-etal-2024-exciting, tseriotou-etal-2024-tempoformer, wang-etal-2025-end}, including studies investigating LLM-based approaches for individual well-being assessment \citep{ravenda-etal-2025-evidence, wang2025posts, ali-etal-2026-overview}, aligning with the focus of our work. Beyond prediction, LLMs offer great potential in mental health analysis through their ability to generate explanations \citep{{59-yang2023towards,yang2023mentalllama,75-bao2024explainable}}, improving interpretability and diagnostic efficiency. Some studies further use LLMs as feature extractors to identify related factors that enhance downstream performance and explainability \citep{lan-etal-2025-depression,ahmed2025explainable,hills-etal-2026-multistream}. Unlike these studies, we systematically investigate diverse automated annotation strategies and evaluate whether and how enrichment features benefit downstream tasks in a training-free setting.

\paragraph{Online Safety}
A wide range of research has examined different aspects of online safety, such as online harassment \citep{padhi2025echoesharassment}, hate speech \citep{rawat2024hatespeech}, and threats \citep{shah2025threats}. Sharenting, one of our interests, is the practice of parents publicly sharing information about their children on social media, such as photographs, health-related details, or participation in school \citep{tosuntacs2024sharenting}. Although it has attracted substantial media and academic attention \citep{ugwudike2024sharenting,roth2025sharenting}, computational research remains limited. The most closely related work is \citet{schirmer2025detecting}, which
evaluates various language models for detecting implicit objectification
of children in comments on TikTok videos of children largely posted by parents. To the best of our knowledge, we are the first to define and provide benchmarks for the following two novel tasks: sharenting risky behaviours (classification of sharenting types) and sharenting risk (assessment of sharenting risk levels).

\section{Datasets}
\label{sec:Datasets for Enrichment}
We focus on downstream tasks in the mental health and online safety domains. Our hypothesis is that such tasks are dynamic and closely tied to an individual's context, while the number and diversity of underlying multi-label indicators make efficient expert annotation challenging. Thus, LLM-based data enrichment could be a promising direction.

\subsection{Well-being Score Prediction} 
Our downstream mental health task uses the CLPsych 2025 shared task dataset on well-being score prediction ~\citep{tseriotou-etal-2025-overview}, which has 437 Reddit posts each assigned a score from 1 to 10 based on the Global Assessment of Functioning (GAF) scale reflecting social, occupational, and psychological functioning. Two related indicators, life events and symptoms, are then manually annotated for a subset of 350 posts, which we use as ground truth to evaluate the quality of automated data enrichment methods.

\paragraph{Life Events} are recognised as important indicators of mental health conditions~\citep{MentalHealthSurveillance,chen-etal-2024-mapping,lv2025tracking}. They refer to experiences that often lead to significant upheaval in an individual’s life (e.g., marriage or job transitions)~\citep{12majorlifeevents}, although their definition and scope vary depending on the research focus~\citep{li2014majorcongratulations,diaz2024lifeeventfinancialcase}. In this work, we draw inspiration from the life-events taxonomy of \citet{12majorlifeevents} and adapt it to our setting. A team of three annotators and one domain expert, providing advice, annotates posts with 21 fine-grained labels allowing multiple labels per post. Due to limited instances in some classes, these are consolidated into eight categories \textit{(Mental Health, Physical Health, Abuse \& Addiction, Relationship \& Loss, Career \& Education, Financial \& Legal \& Societal, Lifestyle \& Identity \& Environment, None)} for evaluation purposes, following recommendations on maintaining sufficient sample sizes per category for validating automatic annotation performance ~\citep{pangakis2023automatedannotationgenerativeai,törnberg2024bestpracticestextannotation}. Four posts are manually selected as few-shot examples for the eight coarse-grained labels, and the remaining posts are used for evaluation.

\paragraph{Symptoms} are strongly related to well-being prediction \citep{tseriotou-etal-2025-overview}. Inspired by the subfactors of the Hierarchical Taxonomy of Psychopathology (HiTOP)~\citep{kotov2017hierarchical}, we develop 11 symptom labels adapted to our study. Two annotators, with guidance from a domain expert, annotate the posts with multiple labels. As some labels contain only a small number of instances, they are merged solely for assessing automatic annotation performance, yielding seven symptom groups: \textit{Fear and Distress, Suicidal Thoughts, Substance Abuse, Antisocial or Antagonistic Externalizing Behaviour, Detachment, Others (combining Somatoform, Eating Pathology, Sexual, and Thought-disorder-related symptoms), and None.} Five posts are manually selected as few-shot examples for the seven coarse-grained labels, and the remaining posts are used for evaluation.

\subsection{{Sharenting Risk Prediction}}
Our downstream online safety task uses a new sharenting risk dataset, presented here for the first time as a computational resource for the sharenting domain, which involves situations where an individual shares personal or sensitive information about a child. The dataset consists of 1,901 Facebook posts selected from public Facebook groups where sharenting is likely to occur, such as parenting, legal, and health support groups. Each post is assigned a risk level based on the information disclosed about a specific child: high risk (A), moderate risk (B), low risk (C), or no risk (D).

\paragraph{Risky Behaviours} include disclosing a child's information such as age or medical conditions, which are closely related to sharenting risk. We define five broad categories of risky behaviour: \textit{Personal Data; Physical, Mental, or Emotional Health; Intervention Services; Disruptive Home Life; Other/None}. A team of three domain experts further annotates these five risky behaviours, allowing multiple labels per post, for the same sharenting risk dataset across all posts. We select seven posts as few-shot demonstrations for the five behaviour labels, and sample 381 posts as the ground-truth evaluation set to assess the quality of automated data enrichment methods. See more details of the dataset, few-shot demonstration selection, annotation process and definitions in Appendix \ref{Appendix:definition and annotation}.

\section{CFD: Confidence-Aware Fine-Grained Debate Framework}
\label{sec:CFD}
\begin{figure*}[htbp]
    \centering
    \includegraphics[width=\textwidth]{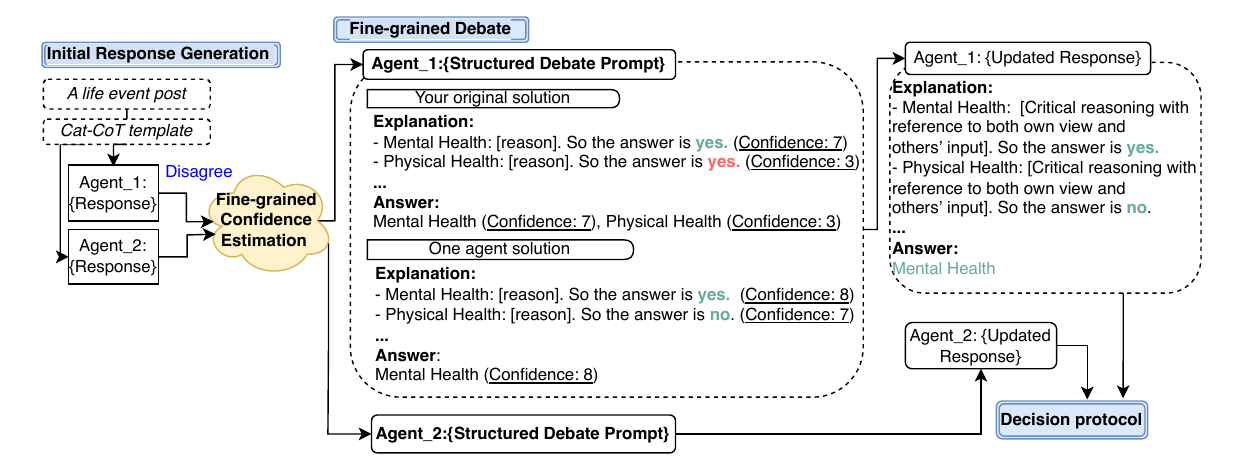}
    \caption{Overview of the proposed CFD framework. Agents first generate independent annotations using Cat-CoT prompting (Section~\ref{subsection:Initial response generation}). When agents disagree, a fine-grained debate is triggered: confidence is estimated at a finer granularity and attached to all agents' responses to guide revision (Section~\ref{subsection:Fine-grained Debate}). Each agent has access to the full conversation history during the debate. Final labels are then determined by the decision protocol (Section~\ref{paragraph:Decision}).}
    \label{fig:illustration_of_CFD_framework}
\end{figure*}

The CFD framework simulates a common collaborative human annotation process for automated data enrichment (Figure~\ref{fig:illustration_of_CFD_framework}): agents first annotate independently (Section~\ref{subsection:Initial response generation}); if they agree, the labels are accepted; otherwise, a fine-grained debate phase is triggered (Section~\ref{subsection:Fine-grained Debate}).

\subsection{Initial Response Generation}
\label{subsection:Initial response generation}
Due to the multi-label nature of many indicators, decomposing the task into multiple single-label predictions is not cost-efficient, whereas standard CoT prompting \citep{weicot2022}, which produces a single reasoning block for all categories, may lead to label omission. To mitigate this, we propose a \textbf{Categorical Chain of Thought (Cat-CoT)} strategy, which requires the model to reason over each predefined category $\mathcal{C} = \{c_1, \dots, c_K\}$ individually, providing both reasoning and a binary judgement (yes/no) for each label. An example of this \textit{category-by-category} output format is shown below, with the full prompt in Appendix~\ref{Appendix: all prompts}.


\begin{tcolorbox}[colback=gray!5,colframe=gray!40,boxrule=0.4pt,left=4pt,right=4pt,top=3pt,bottom=3pt]
{\footnotesize\ttfamily
Explanation:\\
- Mental Health: [reason]. So the answer is yes.\\
- Physical Health: [reason]. So the answer is yes.\\
...\\
- None: [reason]. So the answer is no.\\[2pt]
Answer: Mental Health, Physical Health\par}
\end{tcolorbox}

\subsection{Fine-grained Debate}\label{subsection:Fine-grained Debate}
\paragraph{Structured Debate}
\label{paragraph:Structured Debate}
During the debate stage, each agent receives its own initial output alongside the peer responses, each accompanied by its estimated confidence scores, enabling more precise comparison and revision. For each category, the agent is guided through three steps: reflecting on its own analysis, critically evaluating the peer explanations, and providing comprehensive reasoning on whether to \textit{retain} or \textit{adjust} its original judgement (detailed prompt template in Appendix~\ref{appendix:structured debate}).

\paragraph{Fine-Grained Confidence}
Rather than assigning a single scalar score to an entire response~\citep{yoffe2025debuncimprovinglargelanguage, chen2024reconcileroundtableconferenceimproves}, we estimate fine-grained confidence within each response: a category-level score for each category's reasoning, and an answer-level score for each final predicted label. We evaluate three estimation strategies: \textbf{Self-Verbalised}, \textbf{Entropy-Based}, and \textbf{Sampling-Based} confidence. Once computed, confidence scores are included in the structured debate prompts.

\textbf{(1) Self-Verbalised:}
The self-verbalised confidence method, widely adopted in prior work~\citep{tian2023justaskcalibrationstrategies,chen2024reconcileroundtableconferenceimproves}, leverages the language model's inherent self-assessment capability by prompting the model to explicitly express its confidence. We adopt a two-stage elicitation: once the model has produced its answer, the same model is asked in a separate turn to assign a confidence score between 1 and 10 to its previous response, covering both category-level reasoning and each final selected answer. Confidence is therefore obtained post hoc, as with the other two estimators (prompt provided in Appendix \ref{appendix:Self-verbalised Confidence}).

\textbf{(2) Entropy-Based:}
The entropy-based confidence method derives uncertainty directly from the model's token-level predictive distribution. \citet{yoffe2025debuncimprovinglargelanguage} average the per-token entropy over an entire generation into a single scalar per agent; we instead apply this mean-token-entropy estimator~\citep{fomicheva2020unsupervised} at a finer granularity. Since our output enumerates one rationale per category, we average the per-token entropy within each of them, yielding a per-category uncertainty $u_{i,k}$ for agent $i$ and category $k$. We then convert these uncertainties into confidence scores by adapting their scaling \citep{yoffe2025debuncimprovinglargelanguage}: uncertainties are inverted and scaled so that the mean confidence is $5$ on a $[1,10]$ scale, but we normalise over all $(i,k)$ pairs instead of over one scalar per agent (formulation in Appendix~\ref{Appendix:Entropy-Based Confidence}). The confidence of each selected answer directly reuses its per-category score.\footnote{The answer line typically restates the judgement already made in the explanation, so its entropy is near zero.}

\textbf{(3) Sampling-Based:}
Consistency across multiple sampled generations indicates stable reasoning, whereas higher variability reflects greater uncertainty~\citep{tanneru2023quantifyinguncertaintynaturallanguage}. We therefore sample five additional responses per input, treating the response used in the debate as the reference. At the \emph{category level}, we adapt the chain-of-thought agreement metric of \citet{tanneru2023quantifyinguncertaintynaturallanguage} to a finer granularity: instead of comparing whole explanations, we segment the reasoning of each category into steps and measure the bidirectional entailment agreement between the reference and each sampled reasoning trace using a pre-trained NLI model~\citep{sileo-2024-tasksource}, taking the mean across samples as the confidence of that category's reasoning. At the \emph{answer level}, we additionally estimate the confidence of each final predicted label as the fraction of sampled responses in which it is also predicted. Both scores are linearly scaled to the $1$--$10$ range, and full details are given in Appendix~\ref{Appendix:Sampling-Based Confidence}.

\paragraph{Decision protocol}
\label{paragraph:Decision}
If consensus is reached, the resulting labels are adopted as the final annotations. Otherwise, ties are broken either uniformly at random or by an additional LLM judge, which reviews all rounds of responses from all agents with their confidence scores and selects the final answer (prompts in Appendix~\ref{appendix:LLM-as-a-judge}). To mitigate position bias in the LLM judge, we randomly swap the order of the agents' responses~\citep{zheng2023judgingllmasajudgemtbenchchatbot}.

\section{Automated Data Enrichment}\label{sec:Automated Data Enrichment}
We systematically evaluate a range of automated annotation methods, including CFD, on the life event, symptom, and sharenting risky behaviour datasets described in Section~\ref{sec:Datasets for Enrichment}. These methods differ in how predictions from one or more models are aggregated, and in how uncertainty is represented.

\subsection{Experiments}\label{subsection:Experiments}
\paragraph{Models}
Medium-scale open-weight instruction-tuned LLMs are used for the main evaluation: Qwen2.5-32B-Instruct~\citep{qwen2025qwen25technicalreport} and Mistral-Small-24B-Instruct-2501.\footnote{\url{https://huggingface.co/mistralai/Mistral-Small-24B-Instruct-2501}} These models are chosen for their strong general-purpose capabilities, comparable to larger-scale models while remaining computationally feasible for multi-agent settings\footnote{Despite size differences, their overall abilities are comparable, making them suitable peers in a multi-agent setting.}. We employ Llama-3.3-70B-Instruct~\citep{grattafiori2024llama3herdmodels} as the judge, a larger model from a different family to reduce self-preference bias~\citep{panickssery2024llm}. DeepSeek-R1-Distill-Qwen-32B~\citep{Guo2025Deepseek-r1:Learning} is additionally included as a reasoning-oriented single-LLM baseline. Given the imbalanced label distribution, we adopt Macro-F1 as the primary metric unless otherwise specified. See implementation details in Appendix~\ref{Appendix: experiment details}. 

\paragraph{Initial Generation Format}
We preliminarily compare several initial prompting formats on life events under the zero-shot setting: \textit{Single-label standard CoT}, which queries the model once per label with a single-label prompt and aggregates the binary decisions; \textit{Multi-label standard CoT}, which predicts all labels in a single response; and \textit{Cat-CoT}. Few-shot Cat-CoT is also evaluated. Greedy decoding is used here to ensure a straightforward comparison and stable output formatting.

\paragraph{Evaluated Methods}
We evaluate single-LLM and multi-LLM variants. For multi-LLM variants, we consider random and judge-based tie-breaking following the decision protocol in Section~\ref{paragraph:Decision}. For a fair comparison, all debate-based methods follow a unified process: few-shot Cat-CoT initial generation followed by one interaction round, with \textit{Mistral} and \textit{Qwen} as heterogeneous agents. Debate variants without confidence modelling do not provide confidence signals to the judge.

\textbf{\textit{Single-LLM}}: (1) \textit{Few-shot Cat-CoT}; (2) \textit{Few-shot Cat-CoT + Self-consistency}~\citep{wang2022self}, sampling five reasoning paths and selecting the majority answer. \textbf{\textit{Multi-LLM Ensemble}}: Initial predictions from multiple LLMs are ensembled directly via the decision protocol. \textbf{\textit{Multi-LLM Debate without Confidence}}: \textit{Standard} Debate~\citep{du2023improvingfactualityreasoninglanguage}, where each agent updates its response based on peer responses, without confidence injection. \textbf{\textit{Multi-LLM Debate with Coarse-grained Confidence}}: (1) \textit{DebUnc}~\citep{yoffe2025debuncimprovinglargelanguage}\footnote{ReConcile~\citep{chen2024reconcileroundtableconferenceimproves} is not directly compared, as its convincingness samples require human explanations correcting the model's initial errors, which is not feasible in our low-resource setting.}: a \textit{Standard} Debate variant with response-level confidence estimated via mean token entropy; we adopt their prompt-based confidence injection strategy. (2) \textit{Coarse ablations}: three adapted CFD variants using our structured debate prompt, with a single overall confidence obtained from Entropy (mean token entropy over the whole generation), Verbalised (a single self-reported score), or Sampling (averaging the category-level and answer-level confidences separately, then taking their unweighted mean). \textbf{\textit{Multi-LLM Debate with Fine-grained Confidence}}: CFD instantiated with the above three estimators at the fine-grained level.

\subsection{Results}
\begin{table}[ht]
\centering
\setlength{\tabcolsep}{5pt}
\renewcommand{\arraystretch}{1}
\resizebox{0.44\textwidth}{!}{
\begin{tabular}{@{}llcc@{}}
\toprule
 & & \multicolumn{2}{c}{\textbf{Macro-F1$\uparrow$}} \\
\cmidrule(l){3-4}
\textbf{Setting} & \textbf{Method} & \textbf{Qwen} & \textbf{Mistral} \\
\midrule
Single-label & Std CoT (0-shot) & .622 & .617 \\
\midrule
\multirow{3}{*}{Multi-label}
 & Std CoT (0-shot)  & .660 & .635 \\
 & Cat-CoT (0-shot)  & .692 & .653 \\
 & Cat-CoT (4-shot)  & \textbf{.727} & \textbf{.708} \\
\bottomrule
\end{tabular}
}
\caption{Standard CoT vs. Cat-CoT on Life Events.}
\label{tab:standard-CoTvsCat-CoT}
\end{table}

\begin{table*}[t]
\centering
\setlength{\tabcolsep}{3pt}
\renewcommand{\arraystretch}{1}
\resizebox{0.85\textwidth}{!}{%
\begin{tabular}{@{}>{\raggedright\arraybackslash}p{5.5em} >{\raggedright\arraybackslash}p{7em} l cccccc@{}}
\toprule
 & & & \multicolumn{2}{c}{\textbf{Life Events}} & \multicolumn{2}{c}{\textbf{Symptoms}} & \multicolumn{2}{c}{\textbf{Risky Behav.}} \\
\cmidrule(lr){4-5} \cmidrule(lr){6-7} \cmidrule(lr){8-9}
\multicolumn{3}{@{}l}{\textbf{Setting}} & All & Cont. & All & Cont. & All & Cont. \\
\midrule
\multirow[t]{4}{=}{Single LLM (Cat-CoT)}
 & \multicolumn{2}{l}{Qwen}                     & .737$_{\pm.021}$ & -- & .675$_{\pm.015}$ & -- & .808$_{\pm.003}$ & -- \\
 & \multicolumn{2}{l}{Mistral}                  & .715$_{\pm.014}$ & -- & .658$_{\pm.025}$ & -- & .799$_{\pm.013}$ & -- \\
 & \multicolumn{2}{l}{Llama}                    & .704$_{\pm.022}$ & -- & .662$_{\pm.009}$ & -- & .792$_{\pm.004}$ & -- \\
 & \multicolumn{2}{l}{DeepSeek-R1-Distill-Qwen} & .721$_{\pm.009}$ & -- & .683$_{\pm.004}$ & -- & .785$_{\pm.008}$ & -- \\
\midrule
\multirow[t]{2}{=}{Single LLM (SC)}
 & \multicolumn{2}{l}{Qwen}    & .735$_{\pm.007}$ & -- & .676$_{\pm.009}$ & -- & \textbf{.814}$_{\pm.011}$ & -- \\
 & \multicolumn{2}{l}{Mistral} & .719$_{\pm.008}$ & -- & .687$_{\pm.010}$ & -- & .805$_{\pm.007}$ & -- \\
\midrule
\multirow[t]{2}{=}{Multi-LLM Ensemble}
 & \multicolumn{2}{l}{Qwen \& Mistral (R)} & .721$_{\pm.013}$ & -- & .656$_{\pm.008}$ & -- & .805$_{\pm.009}$ & -- \\
 & \multicolumn{2}{l}{Qwen \& Mistral (J)} & .727$_{\pm.017}$ & -- & .682$_{\pm.011}$ & -- & .811$_{\pm.002}$ & -- \\
\midrule
\multirow[t]{16}{=}{Multi-LLM Debate}
 & \multirow[t]{2}{=}{w/o conf.}
   & Standard (R)   & .716$_{\pm.007}$ & .613$_{\pm.014}$ & .658$_{\pm.029}$ & .538$_{\pm.031}$ & .804$_{\pm.006}$ & .675$_{\pm.012}$ \\
 & & Standard (J)   & .730$_{\pm.011}$ & .629$_{\pm.023}$ & .661$_{\pm.026}$ & .539$_{\pm.029}$ & .812$_{\pm.005}$ & .697$_{\pm.013}$ \\
\cmidrule(l){2-9}
 & \multirow[t]{8}{=}{coarse-grained conf.}
   & DebUnc (R)     & .715$_{\pm.006}$ & .584$_{\pm.016}$ & .649$_{\pm.016}$ & .491$_{\pm.005}$ & .804$_{\pm.008}$ & .656$_{\pm.012}$ \\
 & & DebUnc (J)     & .722$_{\pm.007}$ & .606$_{\pm.007}$ & .667$_{\pm.016}$ & .537$_{\pm.012}$ & .808$_{\pm.010}$ & .681$_{\pm.024}$ \\
 & & Entropy (R)    & .729$_{\pm.021}$ & .635$_{\pm.038}$ & .663$_{\pm.009}$ & .535$_{\pm.023}$ & .798$_{\pm.002}$ & .640$_{\pm.017}$ \\
 & & Entropy (J)    & .733$_{\pm.011}$ & .641$_{\pm.023}$ & .660$_{\pm.016}$ & .518$_{\pm.029}$ & .809$_{\pm.003}$ & .684$_{\pm.002}$ \\
 & & Verbalised (R) & .722$_{\pm.007}$ & .627$_{\pm.015}$ & .669$_{\pm.017}$ & .548$_{\pm.044}$ & .810$_{\pm.004}$ & .682$_{\pm.032}$ \\
 & & Verbalised (J) & .737$_{\pm.008}$ & .645$_{\pm.019}$ & .679$_{\pm.008}$ & .569$_{\pm.032}$ & .808$_{\pm.006}$ & .679$_{\pm.033}$ \\
 & & Sampling (R)   & .733$_{\pm.010}$ & .645$_{\pm.014}$ & .671$_{\pm.020}$ & .542$_{\pm.026}$ & .805$_{\pm.011}$ & .666$_{\pm.046}$ \\
 & & Sampling (J)   & .731$_{\pm.005}$ & .636$_{\pm.001}$ & .670$_{\pm.009}$ & .542$_{\pm.020}$ & .807$_{\pm.007}$ & .662$_{\pm.022}$ \\
\cmidrule(l){2-9}

 & \multirow[t]{6}{=}{fine-grained conf. (ours)}
   & Entropy (R)    & .742$_{\pm.008}$ & .669$_{\pm.014}$ & .665$_{\pm.021}$ & .541$_{\pm.023}$ & .812$_{\pm.019}$ & .694$_{\pm.053}$ \\
 & & Entropy (J)    & \textbf{.748}$_{\pm.016}$ & \textbf{.682}$_{\pm.032}$ & .662$_{\pm.015}$ & .523$_{\pm.011}$ & .807$_{\pm.010}$ & .672$_{\pm.036}$ \\
 & & Verbalised (R) & .730$_{\pm.013}$ & .631$_{\pm.015}$ & .679$_{\pm.002}$ & .561$_{\pm.042}$ & .806$_{\pm.004}$ & .673$_{\pm.010}$ \\
 & & Verbalised (J) & .736$_{\pm.009}$ & .647$_{\pm.007}$ & .680$_{\pm.007}$ & .562$_{\pm.023}$ & .809$_{\pm.005}$ & .681$_{\pm.022}$ \\
 & & Sampling (R)   & .740$_{\pm.010}$ & .655$_{\pm.016}$ & .685$_{\pm.009}$ & .581$_{\pm.030}$ & \textbf{.814}$_{\pm.009}$ & \textbf{.699}$_{\pm.013}$ \\
 & & Sampling (J)   & .746$_{\pm.005}$ & .677$_{\pm.015}$ & \textbf{.702}$_{\pm.011}$ & \textbf{.615}$_{\pm.042}$ & .811$_{\pm.005}$ & .692$_{\pm.033}$ \\
\bottomrule
\end{tabular}
}
\caption{Performance comparison across settings. Macro-F1 ($\uparrow$) is reported as mean $\pm$ standard deviation over 3 runs. (R) and (J) denote random and LLM-as-a-judge tie-breaking, respectively. SC = self-consistency. \textbf{All} is the full evaluation set; \textbf{Cont.} is the contested subset, i.e.\ examples on which agents initially disagree.}
\label{tab:overall_annotation_performance}
\end{table*}

\paragraph{Cat-CoT outperforms standard CoT.}
Table~\ref{tab:standard-CoTvsCat-CoT} compares initial prompting formats on life events, where Cat-CoT consistently outperforms standard CoT on Macro-F1 across models. Interestingly, single-label prompting performs worst despite requiring one call per label. The few-shot setting further improves over the zero-shot setting. We therefore adopt few-shot Cat-CoT as the initial generation format for the main experiments.

\paragraph{Fine-grained confidence debate yields the most robust performance across tasks.}
Table~\ref{tab:overall_annotation_performance} shows our main results. Besides the full evaluation set (\textbf{All}), we also report results on the \emph{contested subset} (\textbf{Cont.}) for the debate methods, i.e.\ the examples on which the two agents initially disagree (26\% on average), since debate can only alter predictions on these examples. We find that fine-grained confidence debate obtains the best performance on Life events and Symptoms and reaches comparable performance to self-consistency on Risky behaviours, with sampling-based CFD performing most consistently across tasks: it significantly outperforms standard debate and DebUnc in 10 of 12 comparisons on All ($p < 0.05$).\footnote{Significance is assessed with two-sided paired approximate randomisation tests (10,000 permutations) on Macro-F1 averaged over the three runs.} Fine-grained confidence debate is also the only multi-LLM setting that exceeds the strongest single-LLM baseline on Life events and Symptoms, which none of standard debate, direct ensemble, or DebUnc achieves, although these gains are not statistically significant. Interestingly, the reasoning model (DeepSeek-R1-Distill-Qwen) does not consistently outperform other single-LLM Cat-CoT baselines.

\begin{figure}[tbht]
\centering
\includegraphics[width=\columnwidth]{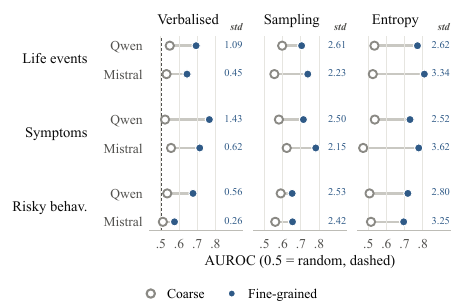}
\caption{Error-detection AUROC of each confidence estimator under coarse and fine-grained confidence. The \textit{std} column gives the within-post standard deviation of label-level scores (fine-grained only).}
\label{fig:confidence_auroc}
\end{figure}

\paragraph{Confidence quality contributes to the benefits of fine-grained debate.}
To examine the effect of confidence, we measure how well each confidence signal distinguishes correct from incorrect decisions on the contested subset. Following \citet{kuhn2023semanticuncertaintylinguisticinvariances}, we report AUROC, the probability that a randomly chosen incorrect decision receives a higher uncertainty score than a randomly chosen correct one, with $1$ indicating perfect separation and $0.5$ a random signal. We compute AUROC per label using its reasoning confidence score, and average across labels and random seeds. Under the coarse setting, a single response-level score is broadcast to all labels, reflecting how an agent may interpret a single confidence value attached to a multi-label response. As shown in Figure~\ref{fig:confidence_auroc}, fine-grained confidence separates correct from incorrect decisions considerably better than coarse confidence does. Among tasks, Risky behaviours shows the weakest separation, which may partly explain the smaller gain of fine-grained confidence debate here. 
\paragraph{Differentiability across labels accounts for the benefits, too.} Fine-grained confidence debate outperforms its coarse counterpart most clearly for entropy and sampling, with statistically significant gains in several settings and up to 7.3 Macro-F1 points for sampling-based CFD on the contested subset (Table~\ref{tab:overall_annotation_performance}). In contrast, the verbalised variant shows no clear advantage despite a comparable increase in AUROC. We therefore report the within-post standard deviation of fine-grained confidence in Figure~\ref{fig:confidence_auroc}, computed as the standard deviation of category-level confidence scores within each post, averaged over posts. It is far lower for verbalised confidence (0.26--1.43) than for sampling (2.15--2.61) and entropy (2.52--3.62), indicating that verbalised scores vary much less across labels within a post and therefore provide a weaker signal for differentiating between them during the debate. Answer-level confidence might also play a role: the sampling estimator computes it separately, providing a complementary signal, whereas the entropy estimator reuses its category-level score. This may explain why sampling-based CFD delivers the most consistent results across tasks.\footnote{We find that removing answer-level confidence from the sampling-based CFD leads to a slight drop in performance.}


\section{Enrichment for Downstream Tasks}
To assess whether and how automated enrichment delivers downstream value, we evaluate several \textit{training-free integration strategies} on two downstream tasks: well-being score and sharenting risk prediction.

\subsection{Experiments}
For the well-being task, we evaluate predictions on the 350-post dataset, excluding 10 instances whose well-being labels are unavailable as they belong to the official dataset’s test split \citep{tseriotou-etal-2025-overview}, and the few-shot examples used for the corresponding indicator (Section~\ref{sec:Datasets for Enrichment}). Following \citet{tseriotou-etal-2025-overview}, we report mean squared error (MSE). For sharenting risk prediction, due to the imbalanced label distribution, we use Macro-F1 as the primary evaluation metric.

\paragraph{Baseline}
We adopt a zero-shot chain-of-thought setup as the baseline, where the prompt consists of the task definition, instruction, post, and output format. We use Llama-3.3-70B-Instruct for all downstream experiments, and each is run three times to report the mean and standard deviation.

\paragraph{Enrichment Integration Strategies}
Building on this baseline, we append the enrichment information to the baseline prompt as additional context (i.e., task definition, instruction, post, followed by enrichment and output format), and the strategies differ only in what is appended (prompts in Appendix~\ref{Appendix: all prompts}). Label-based strategies (1, 2, 4--5) append only the final label names, whereas reasoning-based strategies (3, 6--7) append the raw responses without any aggregated labels. We evaluate the best single- and multi-LLM enrichment methods, self-consistency and sampling-based CFD: (1) \textbf{\textit{Ground-truth labels}}: human-annotated label names. (2) \textbf{\textit{SC labels}}: majority-vote label names from the best Cat-CoT self-consistency configuration (e.g., Mistral for symptoms). (3) \textbf{\textit{SC reasoning}}: all five sampled responses with their CoT traces. (4--5) \textbf{\textit{CFD labels (R/J)}}: team-predicted label names from CFD under random or judge tie-breaking. (6--7) \textbf{\textit{CFD transcripts (R/J)}}: all agent responses across rounds from CFD under the same two settings, with the judge response included when applicable.

\begin{table}[t]
\centering
\setlength{\tabcolsep}{3pt}
\renewcommand{\arraystretch}{1}
\resizebox{\columnwidth}{!}{%
\begin{tabular}{@{}l cc c@{}}
\toprule
& \multicolumn{2}{c}{\makecell{\textbf{Well-being}\\(MSE$\downarrow$)}} & \makecell{\textbf{Sharenting}\\(Macro-F1$\uparrow$)} \\
\cmidrule(lr){2-3}\cmidrule(lr){4-4}
\textbf{Integration} & \makecell{Life\\events} & Symptoms & \makecell{Risky\\behav.} \\
\midrule
Post only (baseline)  & 4.15$_{\pm.24}$ & 4.17$_{\pm.25}$ & .561$_{\pm.022}$ \\
\midrule
+ GT labels           & 3.80$_{\pm.17}$ & \textbf{3.38}$_{\pm.04}$ & .571$_{\pm.010}$ \\
+ SC labels           & 3.95$_{\pm.05}$ & 3.52$_{\pm.00}$ & .575$_{\pm.020}$ \\
+ SC reasoning        & \textbf{3.65}$_{\pm.16}$ & 3.75$_{\pm.15}$ & .653$_{\pm.003}$ \\
+ CFD labels (R)      & 3.93$_{\pm.08}$ & 3.54$_{\pm.08}$ & .571$_{\pm.017}$ \\
+ CFD transcripts (R)   & \textbf{3.71}$_{\pm.11}$ & 3.56$_{\pm.02}$ & \textbf{.660}$_{\pm.016}$ \\
+ CFD labels (J)      & 3.94$_{\pm.09}$ & \textbf{3.48}$_{\pm.05}$ & .573$_{\pm.018}$ \\
+ CFD transcripts (J)   & 3.75$_{\pm.07}$ & 3.55$_{\pm.03}$ & \textbf{.655}$_{\pm.022}$ \\
\bottomrule
\end{tabular}}
\caption{Downstream performance with different enrichment integration strategies. GT = ground-truth.}
\label{tab:downstream-integration-results}
\end{table}

\subsection{Results}
As shown in Table~\ref{tab:downstream-integration-results}, the baseline without enrichment performs worst on all downstream tasks, indicating that enrichment consistently benefits downstream prediction. Moreover, LLM-predicted labels perform comparably to ground-truth labels, suggesting that moderate prediction noise has limited impact on downstream performance.

However, whether labels or reasoning traces help more depends on how the enrichment feature relates to the downstream task, and more context is not always better. For symptoms, which are closely tied to well-being, labels are most effective: ground-truth labels perform best, followed by CFD labels (J), the best-performing symptom annotation strategy. Although symptom severity is more informative for well-being than symptom presence, our enrichment captures only presence, so reasoning traces may add little beyond the labels themselves; indeed, using them instead brings no gain and can even hurt (e.g., 3.75 for SC reasoning vs.\ 3.52 for SC labels). In contrast, life events relate to well-being less directly, as the same event can affect individuals very differently, so labels alone carry limited signal. Reasoning traces can retain contextual cues that labels discard, such as the poster's emotional state; accordingly, all reasoning-based strategies outperform label-based ones (3.65--3.75 vs.\ 3.80--3.95 MSE).

For sharenting risk prediction, behaviour labels are closely related to risk but capture only high-level categories, whereas risk levels depend on specific details that can emerge in the reasoning behind behaviour labels. For example, a post containing personal data may correspond to different risk levels depending on the data type (e.g., specific ages for A vs.\ school year groups for C). Accordingly, reasoning-based enrichment yields the largest gains: CFD transcripts (R) improve Macro-F1 by 9.9 points over the baseline, whereas label-based strategies improve it by at most 1.4 points. Overall, the most effective integration format depends on how strongly the enrichment feature relates to the downstream task, and whether its reasoning carries task-relevant information beyond the label.

\section{Failure Analysis}
\begin{table}[tb]
\centering
\small
\setlength{\tabcolsep}{3pt}
\resizebox{\columnwidth}{!}{%
\begin{tabular}{lcccc}
\toprule
 & \multicolumn{2}{c}{Debate changes} & \multicolumn{2}{c}{Confidence} \\
\cmidrule(lr){2-3}\cmidrule(lr){4-5}
Task & \makecell{C$\rightarrow$I\\($\downarrow$)} & \makecell{I$\rightarrow$C\\($\uparrow$)} &
\begin{tabular}[c]{@{}c@{}}Peer conf.\\ higher (\%)\end{tabular} &
\makecell{AUROC$_{\mathrm{own}}$\\($\uparrow$)} \\
\midrule
Life events  & 15.4 & 20.4 & 57.7 & .623 \\
Symptoms     & 14.4 & 17.8 & 46.5 & .584 \\
Risky behav. & 15.5 & 19.4 & 61.9 & .665 \\
\bottomrule
\end{tabular}
}
\caption{Debate changes of CFD-Sampling on disagreed decisions. C$\rightarrow$I / I$\rightarrow$C: \% of disagreed decisions changed from correct to incorrect / incorrect to correct. Peer conf.\ higher: \% of C$\rightarrow$I changes where the peer's initial confidence on that label exceeded the agent's own. AUROC$_{\mathrm{own}}$: probability that a correct decision an agent kept had higher own initial confidence than one it changed to incorrect ($0.5$ = random). All values are means over 3 seeds.}
\label{tab:debate_changes}
\end{table}

\paragraph{CFD}
We first analyse how CFD-Sampling changes disagreed decisions, i.e.\ each agent's decision on a label where the two agents initially disagree (Table~\ref{tab:debate_changes}). C$\rightarrow$I, I$\rightarrow$C and Peer conf.\ higher are computed over all labels and agents together, whereas AUROC$_{\mathrm{own}}$ is computed for each agent separately, pairing kept and abandoned correct decisions only within the same label, and then averaged over agents. Debate changes 17.8--20.4\% of these decisions from incorrect to correct, but also 14.4--15.5\% from correct to incorrect.
Correct decisions that agents abandoned tend to carry lower own confidence than those they kept (AUROC$_{\mathrm{own}}$ of .584--.665), and for 46.5--61.9\% of these abandoned decisions, the peer had reported higher confidence. Confidence thus plays a role in how agents revise their decisions. However, since confidence is an imperfect signal of correctness (Figure~\ref{fig:confidence_auroc}), a more confident peer is not necessarily correct. Labels on which agents initially disagree could therefore be prioritised for expert review when CFD is applied in practice to sensitive domains.

\paragraph{Downstream Prediction}
We also analyse how errors of the best-performing configuration for each downstream task (Table~\ref{tab:downstream-integration-results}) are distributed across true score ranges and risk levels. For well-being (Table~\ref{tab:wellbeing_failure}), errors are largest on severely impaired posts (scores 1--4). More broadly, predictions are compressed onto a few levels: the three score ranges are most often predicted as 5, 7 and 9 respectively. For sharenting (Appendix Table \ref{tab:sharenting_failure}), high-risk posts (A) have the lowest recall (58.5\%), and across all risk levels, the most frequent error is predicting (A) as low risk (C). Inspecting the model's responses, we find that it can apply a stricter threshold than the definitions require. For well-being, the model notes cues such as daily self-harm urges or inability to maintain personal hygiene, but matches them to the broad description of score 5 (serious symptoms) and judges that they do not amount to a complete inability to function in all areas. For sharenting, a child's exact age alone already defines high risk, yet the model waits for further sensitive information and assigns low risk instead, whose definition includes partial age or location data. In some cases, posts also lack sufficient temporal context to support a judgement. Besides, our downstream experiments use zero-shot prompting with textual scale definitions only, as their purpose is to isolate the effect of enrichment; few-shot examples of boundary cases could potentially help the model mitigate these errors.

\begin{table}[t]
\centering
\small
\setlength{\tabcolsep}{4pt}
\resizebox{\columnwidth}{!}{%
\begin{tabular}{lcccccc}
\toprule
 & \multicolumn{3}{c}{Life events} & \multicolumn{3}{c}{Symptoms} \\
\cmidrule(lr){2-4}\cmidrule(lr){5-7}
Score & $n$ & MSE$\downarrow$ & Mode & $n$ & MSE$\downarrow$ & Mode \\
\midrule
1--4 (Ser.)  & 74  & 6.78 & 5 & 72  & 6.78 & 5 \\
5--6 (Imp.) & 60  & 2.91 & 7 & 62  & 3.25 & 7 \\
7--10 (Min.) & 202 & 2.72 & 9 & 201 & 2.20 & 9 \\
\bottomrule
\end{tabular}
}
\caption{Well-being prediction errors by the score ranges of \citet{tseriotou-etal-2025-overview}. Ser./Imp./Min.: serious impairment to functioning, impaired functioning and minimal impairment to functioning. $n$: number of posts in the range. Mode: the most frequently predicted score. Values are means over 3 seeds.}
\label{tab:wellbeing_failure}
\end{table}

\section{Conclusion}
\label{sec:Conclusion}
We introduce a confidence-aware fine-grained debate framework for automated multi-label enrichment with open-weight LLMs, together with two new expert-annotated resources for mental health and online safety. Our results show the robust performance of CFD and the informativeness of fine-grained confidence, whose benefit is influenced by its quality and variability across labels. Our downstream evaluation further shows that automated enrichment consistently improves performance: LLM-predicted labels are comparable to human labels, and generated reasoning can further surpass them when it carries useful context beyond the label names human annotation typically provides. Our resources and baselines provide a foundation for future work on data enrichment.

\section{Limitations and Future Work}
\label{sec:Limitations}
We do not use commercial LLMs due to privacy constraints associated with our ethical approval, nor the latest crop of reasoning models in the multi-agent setting due to compute cost. Compute constraints also prevent us from evaluating a wider range of models; instead, we carefully select agents of strong and comparable capability that can run together on a single GPU during debate, and a larger judge from a different model family (Section~\ref{subsection:Experiments}). We also do not compare the LLM judge's decisions with independent human adjudication, which would help assess their alignment with human judgements.

Compute cost is also a consideration for CFD itself. Our implementation builds on an existing debate framework~\citep{yoffe2025debuncimprovinglargelanguage} and could be further optimised to reduce GPU hours. As all methods share the same inference settings, we compare costs by the number of input and output tokens. Relative to the strongest single-LLM baseline, self-consistency, entropy-based CFD uses about 0.56$\times$ the tokens, whereas sampling-based CFD uses about 1.5$\times$, as it requires additional sampled generations. Entropy-based CFD is therefore cheaper, while sampling-based CFD performs more consistently across tasks at a higher cost. Investigating more efficient variants is left for future work.

Finally, our experiments focus on the mental health and online safety domains. While we expect our data enrichment framework to be suitable for many other domains, particularly those involving multi-label indicators, this remains to be confirmed in future work.

\section{Ethical considerations}
\label{sec:ethics}
This study received ethical approval from the research ethics committee of the University of Southampton (ERGO ref 88805.A1). While the data comes from publicly available content on Reddit and Facebook, for the life event and symptom datasets we release only Reddit post identifiers together with their corresponding labels, and not the post text itself. The sharenting dataset can be requested by email (\texttt{sem03\{@\}soton.ac.uk}) for academic research only by institutions that can evidence strong ethical approval processes. Dataset access will be approved on a case-by-case basis following our institution's ethics review process. As the data relates to children and may contain sensitive health or safeguarding information, the dataset will be pseudonymised (names of authors and children, email addresses, phone numbers, locations such as schools etc.) before access is granted to any researcher, and the release terms and conditions will prohibit both re-identification and redistribution, and require secure storage and deletion of the data once their project is complete. We collect only public posts, but recognise that the children described did not choose to be written about; by strictly controlling dataset access we intend to protect the privacy of both the post authors and the children, while still allowing important research on the sharenting domain, which ultimately aims to protect children. The corpora may contain distressing material (e.g., self-harm, suicidal ideation); annotators were informed in advance about the nature of the material and were free to withdraw at any point.

Although our enrichment produces indicators such as life events, symptoms, and risky behaviours, these outputs are not intended for diagnostic or assessment purposes and should not be leveraged as such, particularly without expert oversight. They form supplementary information that can be reviewed by domain experts in the loop, augmenting expert capacity by surfacing relevant information efficiently rather than replacing human judgement. Likewise, neither the framework nor the annotated data should be used to monitor, profile, or make decisions about specific individuals or families. Applying our framework to label sensitive data requires following responsible AI and data protection practices.

\section*{Acknowledgements}
\label{sec:Acknowledgements}
This work was supported by the Economic and Social Research Council (ES/V011278/1) and Engineering and Physical Sciences Research Council (EP/Y009800/1), through funding from Responsible AI UK (KP0016) and by MRC grant (grant no. MR/X030725/1). We acknowledge the IRIDIS High-Performance Computing Facility at the University of Southampton. We also thank Silke Roth and Arpan Mandal (University of Southampton) from the sharenting project team for their domain expertise and contributions to the construction of the sharenting dataset. 



\appendix
\section{Confidence Estimation}
\subsection{Entropy-Based Confidence}
\label{Appendix:Entropy-Based Confidence}
Let $i \in \{1,\dots,n\}$ index agents and $k \in \{1,\dots,K\}$ index categories. For each pair, we compute $u_{i,k}$ as the mean token entropy over the tokens of the reasoning generated for category $k$ by agent $i$.

To convert uncertainty values into confidence scores, we follow the scaling of \citet{yoffe2025debuncimprovinglargelanguage}, who normalise one scalar per agent so that the mean confidence across agents is $5$. We normalise over all $(i,k)$ pairs instead. Let $\mathcal{V}$ denote the set of these pairs and $M = |\mathcal{V}|$. We first invert the uncertainties to obtain unscaled confidence values, and then scale them so that the average confidence over $\mathcal{V}$ equals $5$:

{\small
\begin{equation}
c_{i,k} = \frac{1}{u_{i,k}}, \qquad
s_{i,k} = \frac{c_{i,k}}{\sum_{(j,l) \in \mathcal{V}} c_{j,l}} \cdot (5M - 1) + \frac{1}{M}
\end{equation}
}
Finally, we clamp $s_{i,k}$ to the range $1$ to $10$ and round to the nearest integer.

\subsection{Sampling-Based Confidence}
\label{Appendix:Sampling-Based Confidence}
\textit{\textbf{Category-Level:}} For each predefined category $c \in \mathcal{C}$, let $\mathcal{C} = \{c_1, \dots, c_K\}$ denote the set of $K$ labels. Given the reference explanation block $E^{(0)}$ in the Cat-CoT output format (Section \ref{subsection:Initial response generation}) and $N$ sampled counterparts $\{E^{(i)}\}_{i=1}^N$, each reasoning of category $c$, denoted as $S^{(i)}_c$, is segmented into multiple sub-steps via sentence tokenisation up to the phrase \textit{``so the answer is''}:

{\small
\[
S^{(i)}_c = \left[s^{(i)}_{c,1},\,s^{(i)}_{c,2},\,\dots\right], \quad S^{(i)}_c \in E^{(i)}
\]
}

For each pair of sub-steps $(s^{(0)}_{c,p}, s^{(i)}_{c,q})$, a pre-trained NLI model \footnote{https://huggingface.co/tasksource/ModernBERT-large-nli} is used to assess semantic equivalence:

{
\small
\begin{equation}
\begin{split}
\resizebox{0.9\linewidth}{!}{$
f_{\mathrm{NLI}}(p, q) =
\begin{cases}
1, & \text{if } \mathrm{entailment} \text{ and score } \ge 0.5, \\
0, & \text{otherwise.}
\end{cases}
$}
\end{split}
\end{equation}
}

The agreement score between the reference and sampled reasoning for category $c$ is then computed as Eq. \ref{equation:2}, where $P$ and $Q$ are the numbers of steps in $S^{(0)}_c$ and $S^{(i)}_c$, respectively. 

\begin{equation}
\resizebox{\linewidth}{!}{$
\mathrm{AGR}(S^{(0)}_c, S^{(i)}_c) =
\frac{
\sum_{p=1}^{P} \max_{q} f_{\mathrm{NLI}}(p, q)
+ \sum_{q=1}^{Q} \max_{p} f_{\mathrm{NLI}}(p, q)
}{P + Q}
\label{equation:2}
$}
\end{equation}

The confidence of $S^{(0)}_c$ is then estimated as the mean agreement score across all sampled generations:

{\small
\begin{equation}
Conf^{\mathrm{exp}}(c) = \frac{1}{N}\sum_{i=1}^N \mathrm{AGR}(S^{(0)}_c, S^{(i)}_c)
\label{eq:explanation_per_category}
\end{equation}
}

\textit{\textbf{Answer-Level:}}
To reflect uncertainty in the final answer prediction, let $\mathcal{A}^{(0)}$ be the predicted label set from the reference response and $\{\mathcal{A}^{(i)}\}_{i=1}^N$ those from $N$ sampled generations. For each $c \in \mathcal{A}^{(0)}$, we define its confidence as the fraction of samples in which $c$ is also predicted:

{
\small
\begin{equation}
\resizebox{0.9\linewidth}{!}{$
\operatorname{Conf}^{\text{ans}}(c)
= \frac{1}{N} \sum_{i=1}^{N} \mathbf{1}\{\, c \in \mathcal{A}^{(i)} \,\},
\, c \in \mathcal{A}^{(0)}
$}
\end{equation}
}

\section{Experiment Details}\label{Appendix: experiment details}
\paragraph{Implementation details}
Our implementation builds on DebUnc\footnote{\url{https://github.com/lukeyoffe/debunc}} and uses the Hugging Face Transformers library \citep{wolf2020transformers} to run all models. Unless otherwise specified, we follow the temperature settings recommended in the official model documentation, setting the decoding temperature to 0.7 for Qwen2.5 and 0.6 for Llama3.3 and DeepSeek-R1-Distill-Qwen. For Mistral Small 3, whose documentation recommends a relatively low temperature (around 0.15), we adopt a slightly higher value of 0.3 to improve generation diversity. We set top-$k$ to 20 and top-$p$ to 0.8 for all models. For experimental efficiency and to fit the models of a multi-LLM setting on a single GPU, all models used in our experiments are quantised to 4-bit precision. Inference is run on a single NVIDIA A100 80GB GPU. We estimate the total computational budget for all the main experiments at approximately 425 GPU hours.

\section{Datasets and Annotation Process}\label{Appendix:definition and annotation}
\subsection{Annotation}
The life-event annotations are carried out by three PhD students, all fluent in English, including one native speaker. The life-event annotation labels are designed with the help of a domain expert. For the 21 fine-grained categories used during annotation, the mean Fleiss'~$\kappa$ across labels is 0.67 prior to discussion, indicating a sufficient level of annotator agreement. After consolidating these into coarse-grained categories for evaluation, the mean $\kappa$ increases to 0.71. The symptom annotations are conducted by two Master's students from the Department of Psychology, and labels are designed with the help of a domain expert. Both annotators independently annotate a shared subset of 95 posts, achieving a high inter-annotator agreement of 0.82. The remaining posts are then divided equally between them for independent annotation. For the sharenting behaviour and risk datasets, the label design and annotation are carried out by three domain experts, who annotate disjoint subsets of the data. During label development, experts conduct preliminary screening and discussions to refine the sharenting behaviour and risk annotation guidelines. The resulting guidelines provide detailed criteria for each risk level and illustrative examples for each behaviour label, facilitating the final annotation process, with no major difficulties reported in applying the labels. Across all annotation processes, the annotator teams meet regularly, and disagreements are resolved by a mixture of group discussions followed by consensus or majority voting if there is no consensus. Few-shot demonstrations are manually selected with guidance from annotators to include both representative examples of each category and borderline cases. We prioritise instances with multiple labels to maximise label coverage while ensuring that each label is represented at least once. All chain-of-thought explanations are developed under expert guidance. Tables~\ref{tab:life-event-distribution}, \ref{tab:symptom-distribution} and \ref{tab:sharenting-distribution} present the label distributions of our annotated life event, symptom and sharenting datasets, while Table~\ref{tab:wellbeing-distribution} shows the label distribution of the well-being dataset we use.

\begin{table}[t]
\centering
\small
\setlength{\tabcolsep}{4pt}
\renewcommand{\arraystretch}{1.05}
\begin{tabular}{@{}l r@{}}
\toprule
\textbf{Category} & \textbf{Count} \\
\midrule
\textbf{Mental Health} & \textbf{85} \\
\quad Health [mental health diagnosis/recovery] & 59 \\
\quad Health [suicide attempt] & 5 \\
\quad Health [mental health episode] & 34 \\
\textbf{Physical Health} & \textbf{12} \\
\quad Health [accident/illness] & 5 \\
\quad Health [pregnancy] & 2 \\
\quad Health [hospitalization] & 7 \\
\textbf{Abuse \& Addiction} & \textbf{35} \\
\quad Health [abuse] & 18 \\
\quad Health [addiction] & 17 \\
\textbf{Relationship \& Loss} & \textbf{52} \\
\quad Relationship [family] & 22 \\
\quad Relationship [romantic and other] & 38 \\
\quad Death & 5 \\
\textbf{Career \& Education} & \textbf{27} \\
\quad Career & 19 \\
\quad Education & 10 \\
\textbf{Financial \& Legal \& Societal} & \textbf{22} \\
\quad Financial & 10 \\
\quad Legal & 6 \\
\quad Societal & 8 \\
\textbf{Lifestyle \& Identity \& Environment} & \textbf{27} \\
\quad Lifestyle Change & 18 \\
\quad Identity [gender and sexuality] & 3 \\
\quad Identity [religious or political beliefs] & 2 \\
\quad Relocation & 7 \\
\textbf{None} & \textbf{200} \\
\bottomrule
\end{tabular}
\caption{Distribution of coarse-grained (bold) and fine-grained labels.}
\label{tab:life-event-distribution}
\end{table}

\begin{table}[t]
\centering
\small
\setlength{\tabcolsep}{4pt}
\renewcommand{\arraystretch}{1.05}
\begin{tabular}{@{}l r@{}}
\toprule
\textbf{Category} & \textbf{Count} \\
\midrule
\textbf{Fear and Distress} & \textbf{135} \\
\textbf{Suicidal Thoughts} & \textbf{44} \\
\textbf{Substance Abuse} & \textbf{10} \\
\textbf{Antisocial or Antagonistic Externalizing Behavior} & \textbf{13} \\
\quad Antagonistic Behavior & 9 \\
\quad Antisocial Behavior & 5 \\
\textbf{Detachment} & \textbf{28} \\
\textbf{Others} & \textbf{18} \\
\quad Somatoform & 2 \\
\quad Eating Pathology & 4 \\
\quad Sexual & 6 \\
\quad Thought Disorder & 6 \\
\textbf{None} & \textbf{193} \\
\bottomrule
\end{tabular}
\caption{Distribution of coarse-grained (bold) and fine-grained labels.}
\label{tab:symptom-distribution}
\end{table}

\begin{table}[t]
\centering
\small
\setlength{\tabcolsep}{4pt}
\renewcommand{\arraystretch}{1.05}
\begin{tabular}{@{}l r@{}}
\toprule
\textbf{Category} & \textbf{Count} \\
\midrule
\multicolumn{2}{@{}l}{\textit{Risky behaviours}} \\
\midrule
Personal Data & 671 \\
Physical, Mental, or Emotional Health & 396 \\
Intervention Services & 290 \\
Disruptive Home Life & 460 \\
Other/None & 688 \\
\midrule
\multicolumn{2}{@{}l}{\textit{Risk labels}} \\
\midrule
A & 606 \\
B & 303 \\
C & 428 \\
D & 564 \\
\bottomrule
\end{tabular}
\caption{Distribution of sharenting risky behaviours and risk labels on full data.}
\label{tab:sharenting-distribution}
\end{table}

\begin{table}[t]
\centering
\small
\setlength{\tabcolsep}{4pt}
\renewcommand{\arraystretch}{1.05}
\begin{tabular}{@{}l r@{}}
\toprule
\textbf{Well-being} & \textbf{Count} \\
\midrule
10 & 142 \\
9 & 2 \\
8 & 13 \\
7 & 46 \\
6 & 30 \\
5 & 32 \\
4 & 38 \\
3 & 17 \\
2 & 13 \\
1 & 7 \\
\bottomrule
\end{tabular}
\caption{Distribution of well-being scores.}
\label{tab:wellbeing-distribution}
\end{table}

\subsection{Life Event Label Definitions}
Life events are experiences that have a major personal impact on an individual. They must involve a clearly identifiable occurrence or change. These events may have occurred in the past (explicitly stated, or inferred from context if the impact is clear), be occurring in the present (explicitly described or clearly implied), or be expected in the near future (only if explicitly stated).
\begin{itemize}[itemsep=2pt, parsep=0pt, topsep=4pt, leftmargin=*]
\item \textbf{Legal}: Events involving legal matters or law enforcement that impact the individual's life circumstances or emotional state. This includes violations of the law, arrests, court appearances, legal proceedings, and interactions with the justice system, whether the individual is subject to legal processes or initiating legal actions that have significant personal impact.

\item \textbf{Relocation}: Events where an individual discusses moving to a new place; moving in or out of the family; family members moving into or out of the household; or major travel experiences.

\item \textbf{Education}: Events involving school-related life events (e.g., starting or graduating from school, leaving school without graduating, transfer to a different school, being denied entry to school), important exams, severe 
education challenges, or other educational pursuits.

\item \textbf{Death}: Events involving the death of a person or a pet who was important to the individual.

\item \textbf{Career}: Events involving work-related changes, challenges, or milestones, such as job gains or losses, promotions, demotions, troubles at work, unable to find work, retirement, or becoming a business owner or entrepreneur.

\item \textbf{Financial}: Events involving major financial purchases (e.g. house, car), issues, gains and milestones (e.g., paying off debt), or financial losses (e.g., stolen or damaged property).

\item \textbf{Lifestyle Change}: Events involving significant changes to lifestyle ("way or style of living") habits, focused on adopting new routines that support health and well-being. These may include adjustments in diet; exercise; sleep management; stress management (e.g., through meditation); reducing substance use (e.g. quitting alcohol, drugs, or smoking); fostering positive social connections; getting a new pet; or adopting other physical or mental health practices.

\item \textbf{Societal}: Events with clear personal impact include natural disasters; pandemics; major political events; societal issues; or meeting a celebrity they like.

\item \textbf{Identity [gender and sexuality]}: Events where an individual identifies or changes their gender or sexual orientation, or discusses a new sexual experience.

\item \textbf{Identity [religious or political beliefs]}: Events where an individual explores, adopts, or changes their religious (including religious practices) or political beliefs.

\item \textbf{Health [accident/illness]}: Events relating to accidents or injury (e.g., car accident, physical injury), or serious physical illness have long-term impact (e.g. diagnosis or survival from a major illness). This category also includes chronic physical health conditions and their ongoing management when they are described as having a significant impact on the individual's life.

\item \textbf{Health [pregnancy]}: Experiences related to pregnancy including conception, pregnancy announcements, physical and emotional changes throughout pregnancy, complications, miscarriages, terminations, childbirth, or postpartum experiences.

\item \textbf{Health [hospitalization]}: Events where the individual or someone important to them has been hospitalized, undergone surgery, received emergency medical treatment, or experienced significant medical interventions for physical health conditions. 

\item \textbf{Health [mental Health diagnosis/recovery]}: Discussions about receiving a formal mental health diagnosis, beginning or changing psychiatric medication, starting therapy, or experiencing significant improvement in mental health conditions. This includes pivotal moments in the recovery journey such as joining support groups, successfully implementing coping strategies, or experiencing symptom reduction.

\item \textbf{Health [mental health episode]}: Acute manifestations of mental health conditions that represent significant turning points in an individual's psychological journey. These are often characterised by temporary but visibly intense psychological disturbances. This includes manic episodes, psychotic breaks, severe panic attacks, dissociative episodes, or acute suicidal ideation that necessitates crisis intervention. These episodes represent departures from the individual's baseline functioning and often result in significant disruption to daily life. The category captures instances where mental distress manifests in observable, often alarming behavioural changes requiring immediate attention or intervention.

\item \textbf{Health [suicide attempt]}: Events where an individual attempts to end their own life or engages in self-harming behaviours.

\item \textbf{Health [addiction]}: Experiences related to substance dependency, problematic use patterns, or the recovery process. This encompasses drug or alcohol overdose incidents, acknowledgment of addiction, withdrawal symptoms, relapses, and various stages of recovery. The recovery includes interventions such as rehabilitation programs, twelve-step programs, as well as personal milestones in managing substance use disorders. 

\item \textbf{Health [abuse]}: Events involving experiences of abuse, occurring in either childhood or adulthood. Such abuse may take physical, sexual, emotional, psychological, or other forms.

\item \textbf{Relationship [family]}: Events involving challenges and changes in family dynamics. This includes serious arguments or conflicts with family members, family relationships becoming abusive or toxic, parents' divorce or separation, changes in family structure (e.g. adoption, new children), parenting difficulties, family members' health issues when they significantly impact the poster emotionally. It also covers other major events that affect the family unit or require family adjustment.

\item \textbf{Relationship [romantic and other]}: Events involving relationships outside the family unit with strong emotional impact. This includes the beginning or ending of relationships with significant emotional impact (e.g., close friends or romantic partners), getting married or becoming engaged, divorce or broken engagements, relationships becoming abusive or toxic, serious arguments or challenges within these relationships, and health issues of partners/close friends when they significantly impact the poster emotionally.

\item \textbf{None}: Applies when the post does not clearly reflect any of the above life event categories.
\end{itemize}

\subsection{Symptom Label Definitions}
\begin{itemize}[itemsep=2pt, parsep=0pt, topsep=4pt, leftmargin=*]
\item \textbf{Fear and Distress}: Includes emotional problems like feeling sad, loneliness, hopeless, or worried. This covers depression symptoms (sadness, lack of interest in activities, feelings of worthlessness and guiltiness, decreased or increased appetite, decreased or increased sleep); anxiety (excessive worry, fear, nervousness); stress reactions (feeling overwhelmed or burnt out, emotional exhaustion); and reactions to traumatic events (flashbacks, distressing memories, being easily startled). It also includes obsessions and compulsions;  avoidance of feared situations or trauma reminders; emotional numbing or feeling detached from one's emotions; dissociative experiences (feeling disconnected from oneself or surroundings); and irritability or anger outbursts related to emotional distress.
\item \textbf{Suicidal Thoughts}: Involves thoughts about ending one's life, ranging from passive wishes for death to active planning. This includes passive ideation (wishing to be dead, feeling that life is not worth living); active ideation (considering or planning methods of suicide); preoccupation with thoughts of death; expressing the belief that others would be better off without them (subjective feeling of being a burden); making preparations such as giving away possessions or writing goodbye notes; and seeking means or information related to suicide methods. 
\item \textbf{Somatoform}: Involves physical symptoms that cause significant distress but may not have clear medical explanations. This includes various types of pain (headaches, muscle aches, general pain); unexplained physical sensations; physical fatigue; and physical complaints affecting different body systems. These somatic complaints can also stem from anxiety related to health and body (Illness anxiety disorder).
\item \textbf{Eating Pathology}: Includes problematic cognitions and behaviors related to food, eating patterns, body image, and weight. This includes fear of weight gain (frequent body-checking, obsessive weighing), body image disturbance (negative fixation on weight/shape), self-worth tied to appearance (harsh self-criticism, comparisons), denial of low weight, and guilt/shame around eating (anxiety about food, avoidance of eating situations). Also included is using food as emotional coping and obsessive preoccupation with food, cooking, or weight-related content. Behavioral manifestations include restrictive eating (smaller portions, skipping meals, eliminating food groups, extreme diets, calorie counting) and fasting (extended periods without food to compensate for previous meals), and chewing and spitting (chewing food but not swallowing it). Binge eating involves consuming abnormally large amounts of food with feelings of loss of control, eating rapidly or until uncomfortably full, eating alone due to embarrassment, and experiencing negative emotions afterward. Compensatory behaviors such as self-induced vomiting, misuse of laxatives/diuretics, use of diet pills or "slimming teas," fasting as a non-purging compensatory behavior and excessive exercise (working out despite fatigue/injury, exercise interfering with daily functioning) are also key indicators. This includes secretive behaviors around eating, including attempts to conceal eating patterns or using coded terminology in online communities (e.g., "UW/GW/HW/LW", "wieiad", "Thinspo"). 
\item \textbf{Sexual}: Involves difficulties with sexual functioning and satisfaction. This includes problems with sexual desire, arousal, or performance; pain during sexual activity; and distress related to sexual experiences. Also includes sexual addiction in a way that impairs various other functions. 
\item \textbf{Thought Disorder}: Relates to disruptions in perception, beliefs, and thought organization. This includes unusual sensory experiences (e.g. hearing voices); extreme and fixed, false beliefs; disorganized thinking or speech; and difficulty distinguishing between what's real and what's not.
\item \textbf{Substance Abuse}: Covers problematic use of alcohol, drugs, or other substances. This includes excessive drinking or using recreational drugs; unsuccessful efforts to cut down use; spending significant time obtaining, using, or recovering from substances; and continued use despite negative consequences.
\item \textbf{Antisocial behavior}: Includes actions that violate societal norms and infringe upon the rights of others. This includes breaking rules; physical aggression; blame externalization; disregard for societal norms; deceitfulness or theft; failure to meet obligations; engaging in illegal activities; property destruction; bullying. These behaviors involve little or no remorse. Includes also disinhibited behaviors and symptoms such as impulsivity, distractibility and risk taking (ADHD-related).
\item \textbf{Antagonistic externalizing behavior}: Includes hostile, threatening, or manipulative interactions; extroverted flirts; deliberately annoying others; attention seeking; rudeness; using others for personal (secondary) gain; and other deliberately over-extroverted behaviors (physical or emotional). A feeling of regret typically follows these behaviors. 
\item \textbf{Detachment}: Characterized by withdrawal from social and emotional experiences. The withdrawal may stem from a denial of needing others or from despair. This includes avoiding social interactions; limited emotional expression; lack of close relationships; and reduced ability to experience pleasure in activities (particularly, social anhedonia); isolation.
\item \textbf{None}: Applies when the post does not clearly reflect any of the above symptom categories.
\end{itemize}

\subsection{Risky Behaviour Label Definitions}
Sharenting is when someone shares personal or sensitive information about a specific child (under 18). It also includes posts that might cause harm to the child or to the child's family's reputation. Self-disclosure about oneself, even when under 18, is not considered sharenting.
If the age of the person whose information is being shared is unknown but familial or social roles are mentioned that signal they are a child (e.g., son, daughter, sibling, nephew, niece, grandchild, schoolchild), assume they are children. If they are aged between 18 and 24 (inclusive), sharenting behaviour applies only when they are described in a child role within the family or social context.

1. \textbf{Personal Data}: 
This category includes mentions of a child's GDPR-related personal data, such as:
\begin{itemize}
    \item A child's name (even if partially given), specific age or a specific school year (e.g., Reception, Year 1-11, Sixth Form), and location details such as city, town, or county, but not country. It also includes sensitive attributes such as sexual orientation, race or ethnicity, religious beliefs, or political opinions.
    \item Sharing of images, videos, or audio recordings of a child.
\end{itemize}
2. \textbf{Physical, Mental, or Emotional Health}:
This category includes mentions of a child's physical, mental, or emotional issues, such as: 
\begin{itemize}
    \item Mentions of a child's diagnosed or suspected medical conditions and disorders including ADHD, anxiety, autism, dyslexia, dysgraphia, speech delays, self-harm, or eating disorders.
    \item Seeking official referrals or a diagnosis for a child, or disclosing that a child is attending therapy.
    \item Childhood ailments such as nappy rash, vomiting, toothache, or other aches.
    \item Seeking advice on feeding, weaning, sleep, or potty training.
    \item Discussion of puberty, a child's emotional distress, or a child engaged in negative behaviour such as school refusal, loneliness or not having friends, severe sibling rivalry, violence against siblings or parents, low self-esteem, lying, stealing, bullying, or running away.
    \item Meltdowns or common behaviours in children under five (e.g., tantrums, hitting, biting, screaming).
    \item A child being emotionally upset, for example showing a few tears, which is not serious or chronic distress.
\end{itemize}
3. \textbf{Intervention Services}:
This category covers formal institutional or statutory child-focused programmes or casework where the child is referred into or managed by a public or specialist service, such as:
\begin{itemize}
    \item Social services, social workers, or specialist educational needs and health services, including SENCO, CAMHS, and SENAR.
    \item An EHCP, the SEN register, DLA (Disability Living Allowance), or an IDP (Independent Development Plan).
    \item Serious intervention or safeguarding measures such as Section 47 of the Children Act 1989, a CIN plan (Children in Need), a Child Protection Plan, and the child protection register.
    \item Child criminal behaviour, including criminal allegations, police investigations of a child, or court proceedings against a child (e.g., youth court, juvenile offences).
    \item Child placed in temporary foster care, child 'in care', or child subject to a Special Guardianship Order (SGO).
\end{itemize}
4. \textbf{Disruptive Home Life}:
This category includes mentions of a damaging or disruptive home life of a child that can negatively affect a child's physical, emotional, or mental well-being, or that may cause distress to the child and harm their welfare if publicly known.
It applies where there is an indication of a parental or carer relationship (e.g., mum, dad, custody case) or a reasonable suggestion that the child is living in the home or has home-like contact with the parties at least some of the time.
\begin{itemize}
    \item Domestic violence, non-molestation orders, abuse, or neglect.
    \item Serious disclosures about parents or any adult living with the child (e.g., a sibling aged 25 or over, or a grandparent), including involvement in serious crimes, drug abuse, disclosures of mental health issues, or seriously disparaging a parent's reputation (e.g., accusations of parental alienation behaviours).
    \item A parent saying they cannot cope with the child, a parent in emotional distress because of the child, or family conflict because of the child (for example, the child's misbehaviour leading to parental disagreement).
    \item Refusal to pay child benefit, or a parent rejecting a paternity claim.
    \item Minor breaches of arrangement orders, minor co-parenting conflicts, parental separation or divorce, CAFCASS and family courts, Section 7 of the Children Act 1989, C100 applications, ICFA, child arrangement orders (CAO), or custody of the child.
\end{itemize}
5. \textbf{Other/None}:
Applies when the post does not clearly reflect any of the above sharenting categories. For example, spam, advertisements, news, general non-sensitive sharenting about a child's life (e.g., deregistering from school, favourite toys, activities they do, or things they like and dislike), or general advice-seeking (for example, asking about products, items, or activities for a child that may suggest the child's preferences or aspects of the child's life, such as nightlights or nappies).

\subsection{Sharenting Risk Label Definitions}
A. \textbf{High risk}:
Any explicit disclosure of personal data (as defined by GDPR) about a child or any disclosure that allows inference of health status (e.g., through the disclosure of medication used, treatments engaged in, or involvement with specific intervention teams). Includes:
\begin{itemize}
    \item Specific age (under 18); name (even partial); specific locations (e.g., street name or school name).
    \item Disclosures made about child's sexual orientation, sex life, race or ethnicity, religious ideology, or political opinion.
    \item Sharing of images, videos, or audio recordings of a child.
    \item Child diagnosed with or described as having medical conditions and disorders (ADHD, anxiety, autism, dyslexia, dysgraphia, speech delays, self-harm, eating disorders).
    \item References to a child receiving medical interventions or taking prescribed medication, involvement with specialist educational or health services such as SENCO, CAMHS, or SENAR, or being on an EHCP, SEN register, DLA (Disability Living Allowance), or IDP (Independent Development Plan).
    \item Children as offenders or suspected offenders involved in criminal activity, allegations, investigations, and proceedings.
\end{itemize}
B. \textbf{Moderate risk}:
Disclosure of damaging information that could cause significant damage to the reputation of a child or family, or disclosure of speculative, generic, or non-specific data relating to a child that would otherwise be considered GDPR special category data if it were complete or explicit. Includes:
\begin{itemize}
    \item Assumptions or suspicions about child medical conditions and disorders (not diagnosed). Seeking official referrals or a diagnosis for a child, or disclosing that a child is attending therapy.
    \item References to temporary conditions and common childhood ailments, such as nappy rash, vomiting, toothache, stomach ache, or other aches, as well as discussion of puberty.
    \item References to a child's emotional distress or to a child engaged in serious negative behaviour, such as school refusal, loneliness or not having friends, severe sibling rivalry, violence against siblings or parents, low self-esteem, lying, stealing, bullying, or running away.
    \item Child placed in temporary foster care, child 'in care', or child subject to a Special Guardianship Order (SGO).
    \item Serious disclosures about parents or any adult (e.g., a sibling aged 25 or over, or a grandparent) living with the child for some or part of the time. This includes involvement in serious crimes, domestic violence, non-molestation orders, drug abuse, disclosures of mental health issues, or seriously disparaging a parent's reputation (e.g., 'accusations' of parental alienation or abuse).
    \item References to a child being at risk from serious harm, such as neglect, abuse, usually mentioned alongside serious intervention or safeguarding measures, including Section 47, a CIN plan (Children in Need), a Child Protection Plan, or inclusion on the child protection register, and involvement with social services or social workers.
    \item A parent saying they cannot cope with the child, a parent in emotional distress because of the child, family conflict because of the child (e.g., the child's misbehaviour leading to parental disagreement), refusing to pay child benefit, or a parent rejecting a paternity claim.
\end{itemize}
C. \textbf{Low risk}:
A specific child is mentioned, but no GDPR-protected data is disclosed (except for partial data on age or location). It also includes disclosures about the child or home life that are unlikely to cause significant damage to the reputation of the child or family. Includes:
\begin{itemize}
    \item  Mentions that the person is aged 18-24 (inclusive) and is described in a child role.
    \item Disclosure of school year group that suggests child is in a particular age range.
    \item Location details such as city, town, or county, but not country.
    \item A child being emotionally upset, for example showing a few tears, which is not serious or chronic distress.
    \item Seeking advice on feeding, weaning, sleep, or potty training.
    \item Usual for children, Year 1 and below/just starting school: difficulty making friends, no close friends, occasional reluctance to go to school (if persistent, escalate to moderate risk).
    \item Child engaged in negative behaviour but considered 'minor' rebellious behaviour (e.g., not listening to parents, not doing chores, refusing to go to their room, not doing as told).
    \item Meltdowns or common behaviours in children under five (e.g., tantrums, hitting, lying, biting, screaming).
    \item Minor breaches of arrangement orders, minor co-parenting conflicts, parental separation or divorce, CAFCASS and family courts, Section 7 of the Children Act 1989, C100 applications, ICFA, child arrangement orders (CAO), or custody of the child.
    \item General sharenting about a child's life (e.g., deregistering from school, favourite toys, activities they do, or things they like and dislike), or general advice-seeking (for example, asking about products, items, or activities for a child that may suggest the child's preferences or aspects of the child's life, such as nightlights or nappies).
\end{itemize}
D. \textbf{No risk}:
No sharenting risk about a specific child. For example, posts might be general advice (not related to a specific child), spam, adverts, or news items.

\section{All prompts}\label{Appendix: all prompts}
\subsection{Cat-CoT for life events}
\label{appendix:cat-cot-life-events}
\begin{lstlisting}[basicstyle=\footnotesize,breaklines=true,breakatwhitespace=true, frame=single]
You are provided with a social media post and must identify any personal life events based on the life event categories defined below.

Life events are experiences that have a major personal impact on an individual. They must involve a clearly identifiable occurrence or change. These events may have occurred in the past (explicitly stated, or inferred from context if the impact is clear), be occurring in the present (explicitly described or clearly implied), or be expected in the near future (only if explicitly stated).

- Mental Health: 
Mental health life events cover discussions about receiving a formal mental health diagnosis; starting or adjusting psychiatric medication; beginning therapy; recovery or significant symptom improvement; or experiencing acute psychological episodes such as manic episodes, psychotic breaks, panic attacks, dissociative episodes, self-harm, suicide attempts, or acute suicidal ideation requiring crisis intervention. It does not include general low mood, stress, anxiety, or depression unless explicitly tied to one of these major events.
- Physical Health: 
Physical health life events cover accidents; injuries; diagnoses or survival of serious illnesses; chronic conditions that have a notable impact on daily life; or cases where the individual or someone important to them has been hospitalized, undergone surgery, received emergency medical treatment, or experienced other major medical interventions. Pregnancy-related experiences, including becoming pregnant, pregnancy loss, abortion, or complications, are also included. Common aches, headaches, fatigue, or other symptoms without being part of a major physical health event are not included.
- Abuse & Addiction: 
Covers major life events involving abuse or substance-related issues. Abuse may occur in childhood or adulthood and may take physical, sexual, emotional, psychological, or other forms. Addiction-related events include the onset of heavy drug or alcohol use; acknowledgment of addiction; overdoses; withdrawal symptoms; relapses; or recovery (including participation in treatment programs such as rehabilitation or support groups).
- Relationship & Loss: 
Covers specific relationship changes or disruptions in meaningful personal connections—family or non-family—that result in strong emotional impact. Includes beginning or ending friendships or romantic relationships; marriage or divorce; parental separation or divorce; serious arguments or conflicts; relationships becoming abusive or toxic; the addition of new family members (e.g., through birth or adoption); parenting difficulties; emotionally distressing health issues involving family members, partners, close friends, or other meaningful connections; or the death of a person or pet who was important to the individual.
- Career & Education: 
Career-related life events cover starting or losing a job; promotions; demotions; troubles at work; being unable to find a job; retirement; or becoming a business owner. Education-related life events include starting or graduating from school; transferring schools; leaving school without graduating; being denied entry into school; taking major exams; experiencing significant academic challenges; or achieving major academic milestones (e.g., earning a certification).
- Financial & Legal & Societal: 
Financial life events include major purchases (e.g., a house or car); financial gains or milestones (e.g., paying off debt); financial losses (e.g., stolen or damaged property); or ongoing financial difficulties. Legal life events include law violations; arrests; court appearances; lawsuits or legal actions; or major interactions with the justice system, whether the individual is subject to legal proceedings or initiating them. Societal life events with clear personal impact include natural disasters; pandemics; major political events; societal issues; or notable public encounters (e.g., meeting a celebrity).
- Lifestyle & Identity & Environment: 
Covers major life events involving changes in lifestyle habits, personal identity, or living environment. Lifestyle changes include adopting new health- and well-being-related routines, such as adjustments in diet; exercise; sleep management; stress management (e.g., through meditation); reducing substance use; fostering positive social connections; or getting a pet. It does not include clinical treatment for mental health issues. Identity transitions include identifying or changing one's gender or sexual orientation (e.g., coming out as LGBTQ+); discussing a new sexual experience; or exploring, adopting, or changing political or religious beliefs (including changes in religious practices). Relocation life events include moving to a new place; moving in or out of the family home; family members moving into or out of the household; or significant travel experiences.
- None: 
Applies when the post does not clearly reflect any of the above life event categories.

Instructions:
1. Read the post carefully and evaluate whether it matches each defined life event category.
2. For each life event category, explain your reasoning. If a category applies, support your answer with direct evidence from the post. If it does not apply, explain why there is insufficient or no evidence. Clearly state "yes" or "no" for each category.
3. Finally, list all broad life event categories that apply. If more than one applies, separate them with commas.

Below are some examples:
{few_shot_examples}

Post to Analyze:
Post:
"{post}"

Please strictly follow the output format exactly as shown below. Do not use bold, markdown, or extra formatting.

Output Format:
Explanation:
- Mental Health: [reason]. So the answer is yes (or is no).
- Physical Health: [reason]. So the answer is yes (or is no).
- Abuse & Addiction: [reason]. So the answer is yes (or is no).
- Relationship & Loss: [reason]. So the answer is yes (or is no).
- Career & Education: [reason]. So the answer is yes (or is no).
- Financial & Legal & Societal: [reason]. So the answer is yes (or is no).
- Lifestyle & Identity & Environment: [reason]. So the answer is yes (or is no).
- None: [reason]. So the answer is yes (or is no).

Answer:
Output exact category names: Mental Health, Physical Health, Abuse & Addiction, Relationship & Loss, Career & Education, Financial & Legal & Societal, Lifestyle & Identity & Environment, or None. Use commas to separate multiple labels, if any.
\end{lstlisting}

\subsection{Cat-CoT for symptoms}
\label{appendix:cat-cot-symptoms}
\begin{lstlisting}[basicstyle=\footnotesize,breaklines=true,breakatwhitespace=true, frame=single]
You are provided with a social media post and must identify any current symptoms of psychopathology experienced or expressed by the poster themselves, according to the symptom categories defined below, noting that symptoms may co-occur.

Symptom Categories:
- Fear and Distress:
Includes emotional problems such as feeling sad, loneliness, hopeless, or worried. This covers depressive symptoms (e.g., sadness, lack of interest in activities, feelings of worthlessness or guilt, and changes in appetite or sleep); anxiety symptoms (e.g., excessive worry, fear, nervousness); stress reactions (e.g., feeling overwhelmed or burnt out, emotional exhaustion); and reactions to traumatic events (e.g., flashbacks, intrusive memories, being easily startled).
It also includes obsessions and compulsions (OCD-related); avoidance of feared situations or trauma reminders; emotional numbing or feeling detached from one's emotions; dissociative experiences (feeling disconnected from oneself or surroundings); and irritability or anger outbursts related to emotional distress.
- Suicidal Thoughts:
Distinct from general distress, this involves specific thoughts about ending one's life or engagement in self-harm or suicide-related behaviors. It includes both passive suicidal ideation (e.g., "I wish I wouldn't wake up") and active suicidal ideation (e.g., considering or planning methods of suicide), as well as references to self-harm behaviors (e.g., cutting), or suicide attempts.
- Substance Abuse:
Refers to excessive use of alcohol, drugs, or other substances in a way that is harmful to the individual or others (non-substance addictions, namely behavioral addictions, are not included). This includes excessive drinking or using recreational drugs; unsuccessful efforts to cut down use; spending significant time obtaining, using, or recovering from substances; and continued use despite negative consequences.
- Antisocial or Antagonistic Externalizing Behavior:
Antisocial behavior includes actions that violate societal norms or infringe upon the rights of others, such as breaking rules; physical aggression; blame externalization (e.g., blaming others for one's failures); disregard for societal norms; deceitfulness or theft; failure to meet obligations; engaging in illegal activities; property destruction; bullying. These behaviors involve little or no remorse. It also includes disinhibited behaviors and symptoms such as impulsivity, distractibility, and risk-taking (ADHD-related).
Antagonistic externalizing behavior includes hostile, threatening, or manipulative interactions; extroverted flirts; deliberately provoking or annoying others; attention seeking; rudeness; using others for personal (secondary) gain; and other deliberately over-extroverted behaviors (physical or emotional). A feeling of regret typically follows these behaviors.
- Detachment:
Characterized by withdrawal from social and emotional experiences with others. The withdrawal may stem from a denial of needing others or from despair. This includes avoiding social interactions; limited emotional expression toward others; lack of close relationships; and reduced ability to experience pleasure in social interactions and relationships (i.e., social anhedonia); isolation.
- Others:
Apply this category only when the post describes symptoms that fall into at least one of the following categories: somatoform-related, eating pathology-related, sexual-related, or thought disorder-related symptoms.
Somatoform-related symptoms involve physical symptoms that cause significant distress but may not have clear medical explanations (e.g., unexplained pain such as headaches or muscle aches, unexplained physical sensations, physical fatigue, physical complaints affecting different body systems). These somatic complaints can also stem from anxiety related to health and the body (illness anxiety disorder).
Eating pathology-related symptoms include problematic cognitions and behaviors related to food, eating patterns, body image, and weight (e.g., restrictive eating, binge eating, compensatory behaviors to prevent weight gain, using food as emotional coping, negative fixation on weight or shape).
Sexual-related symptoms involve difficulties with sexual functioning or satisfaction, or sexual addiction that impairs various other functions (e.g., problems with sexual desire, arousal, or performance; pain during sexual activity; distress related to sexual experiences; pornography addiction or excessive masturbation).
Thought disorder-related symptoms relate to disruptions in perception, beliefs, or thought organization (e.g., unusual sensory experiences such as hearing voices, fixed false beliefs, disorganized thinking or speech, or difficulty distinguishing between what's real and what's not).
- None:
Applies when the post does not clearly reflect any of the symptom categories defined above.

Instructions:
1. Read the post carefully and evaluate whether it matches each defined symptom category.
2. For each category, explain your reasoning. If a category applies, support your answer with direct evidence from the post. If it does not apply, explain why there is insufficient or no evidence. Clearly state "yes" or "no" for each category.
3. Finally, list all applicable symptom categories that are present in the post. If more than one applies, separate them with commas.

Below are some examples:
{few_shot_examples}

Post to Analyze:
Post:
"{post}"

Please strictly follow the output format exactly as shown below. Do not use bold, markdown, or extra formatting.

Output Format:
Explanation:
- Fear and Distress: [reason]. So the answer is yes (or is no).
- Suicidal Thoughts: [reason]. So the answer is yes (or is no).
- Substance Abuse: [reason]. So the answer is yes (or is no).
- Antisocial or Antagonistic Externalizing Behavior: [reason]. So the answer is yes (or is no).
- Detachment: [reason]. So the answer is yes (or is no).
- Others: [reason]. So the answer is yes (or is no).
- None: [reason]. So the answer is yes (or is no).

Answer:
Output exact category names: Fear and Distress, Suicidal Thoughts, Substance Abuse, Antisocial or Antagonistic Externalizing Behavior, Detachment, Others, or None. Use commas to separate multiple labels, if any.
\end{lstlisting}

\subsection{Cat-CoT for sharenting risky behaviours}
\label{appendix:cat-cot-risky-behaviours}
\begin{lstlisting}[basicstyle=\footnotesize,breaklines=true,breakatwhitespace=true, frame=single]
You are given a post and must decide whether it contains any of the sharenting behaviours defined below.

Sharenting is when someone shares personal or sensitive information about a specific child (under 18). It also includes posts that might cause harm to the child or to the child's family's reputation. Self-disclosure about oneself, even when under 18, is not considered sharenting.
If the age of the person whose information is being shared is unknown but familial or social roles are mentioned that signal they are a child (e.g., son, daughter, sibling, nephew, niece, grandchild, schoolchild), assume they are children. If they are aged between 18 and 24 (inclusive), sharenting behaviour applies only when they are described in a child role within the family or social context.

Categories:
{Risky Behaviour Label Definitions in C.4}

Constraint: If any of categories 1-4 is "yes", then 5 (Other/None) must be "no". If and only if all of 1-4 are "no", then 5 must be "yes" (and alone).

Instructions:
1. Read the post carefully and evaluate whether it matches each defined sharenting behaviour category.
2. For each category, explain your reasoning. If a category applies, support your answer with direct evidence from the post. If it does not apply, explain why there is insufficient or no evidence. Clearly state "yes" or "no" for each category.
3. Finally, list all categories that apply. If more than one applies, separate them with commas.

Below are some examples:
{few_shot_examples}

Post to Analyse:
Post:
"{post}"

Please strictly follow the output format exactly as shown below. Do not use bold, markdown, or extra formatting.

Output Format:
Explanation:
1. Personal Data: [reason]. So the answer is yes (or is no).
2. Physical, Mental, or Emotional Health: [reason]. So the answer is yes (or is no).
3. Intervention Services: [reason]. So the answer is yes (or is no).
4. Disruptive Home Life: [reason]. So the answer is yes (or is no).
5. Other/None: [reason]. So the answer is yes (or is no).

Answer:
Write only the category numbers. Do not include category names or extra words. Use commas to separate multiple labels, if any.
\end{lstlisting}

\subsection{Structured Debate}
An example structured debate prompt for life events is shown below. Prompts for the other tasks follow the same template with task-specific output formats.
\label{appendix:structured debate}
\begin{lstlisting}[basicstyle=\footnotesize,breaklines=true,breakatwhitespace=true, frame=single]
These are solutions and confidence scores (1 to 10, where higher means more confident) provided by you and the other agents for the given problem. Each category includes an explanation with its own confidence score, and each final selected answer has a single overall confidence score: 

Your original solution:
{your_solution_with_confidence}

One agent solution:
{peer_solution_with_confidence}

Based on your and other agents' opinions and confidence levels, can you provide an updated response?

Before you update your answer, carefully think through the following steps for each category:
1. Reflect on Your Original Analysis: Briefly restate your original reasoning, conclusion, and the key evidence supporting it.
2. Critically Evaluate External Opinions: Analyze the explanations provided by the other agents. Identify any strengths, weaknesses, or potential biases in their reasoning. Point out any evidence or details you think are missing or overemphasized.
3. Synthesize and Update: Considering both your original analysis and the external opinions, provide a comprehensive and objective reasoning that explains whether you should adjust your original conclusion or retain it. Ensure that you discuss both sides before arriving at your final decision. 

Please strictly follow the output format exactly as shown below. Do not output your confidence scores. Do not include bold text, markdown, or additional explanation.

Explanation:
- Mental Health: [Include your critical reasoning with reference to both your view and others' input]. So the answer is yes (or is no).
- Physical Health: [Include your critical reasoning with reference to both your view and others' input]. So the answer is yes (or is no).
- Abuse & Addiction: [Include your critical reasoning with reference to both your view and others' input]. So the answer is yes (or is no).
- Relationship & Loss: [Include your critical reasoning with reference to both your view and others' input]. So the answer is yes (or is no).
- Career & Education: [Include your critical reasoning with reference to both your view and others' input]. So the answer is yes (or is no).
- Financial & Legal & Societal: [Include your critical reasoning with reference to both your view and others' input]. So the answer is yes (or is no).
- Lifestyle & Identity & Environment: [Include your critical reasoning with reference to both your view and others' input]. So the answer is yes (or is no).
- None: [Include your critical reasoning with reference to both your view and others' input]. So the answer is yes (or is no).

Answer:
Output exact category names: Mental Health, Physical Health, Abuse & Addiction, Relationship & Loss, Career & Education, Financial & Legal & Societal, Lifestyle & Identity & Environment, or None. Use commas to separate multiple labels, if any.
\end{lstlisting}

\subsection{Self-verbalised Confidence}
An example self-verbalised confidence elicitation prompt for life events is shown below. Prompts for the other tasks follow the same template with task-specific output formats.
\label{appendix:Self-verbalised Confidence}
\begin{lstlisting}[basicstyle=\footnotesize,breaklines=true,breakatwhitespace=true, frame=single]
Now rate your own confidence in the response you just gave.

Do not revise your answer. Do not restate your reasoning. Do not add any explanation.

Output integer confidence scores from 1 to 10 for each category in the explanation and the final answer, where 1 means completely uncertain and 10 means completely certain, based on how certain you are about your reasoning and conclusion.

Please strictly follow the output format exactly as shown below. Do not use bold, markdown, or extra formatting.

Explanation Confidence:
- Mental Health: X
- Physical Health: X
- Abuse & Addiction: X
- Relationship & Loss: X
- Career & Education: X
- Financial & Legal & Societal: X
- Lifestyle & Identity & Environment: X
- None: X

Answer Confidence:
For each category that appears in your Answer above, in the same order, use the format Category Name: X, and separate multiple categories with commas.
\end{lstlisting}

\subsection{LLM-as-a-Judge}
An example LLM-as-a-judge prompt for life events is shown below. The prompts for symptoms and risky behaviours are the same apart from the task definition and output format. The few-shot examples are the same as in the initial generation prompt.
\label{appendix:LLM-as-a-judge}
\begin{lstlisting}[basicstyle=\footnotesize,breaklines=true,breakatwhitespace=true, frame=single]
You are a judge tasked with evaluating a debate between two agents about whether a given post reflects one or more predefined life event categories.

For each category, analyze and compare the agents' arguments for accuracy and relevance. Judge whether their reasoning is well supported by the post content and whether it aligns with the given category definitions.

When comparing agents, treat confidence scores as secondary signals. Base your decision primarily on the strength and clarity of evidence in the post. High confidence without explicit textual support should be treated as weak reasoning.

Then, based on this comparison, decide whether the category applies and explain your reasoning.

---

{Task_Definition}

---

Below are some examples:
{few_shot_examples}

Post to Analyze:
"{post}"

Transcript of the Debate:

agent_0:
Round 1:
{response_with_confidence}
Round 2:
{response_with_confidence}

agent_1:
Round 1:
{response_with_confidence}
Round 2:
{response_with_confidence}

---

Please follow the exact format below.

Output Format:
Explanation:
- Mental Health: [Compare, explain, and conclude]. So the answer is yes (or is no).
- Physical Health: [Compare, explain, and conclude]. So the answer is yes (or is no).
- Abuse & Addiction: [Compare, explain, and conclude]. So the answer is yes (or is no).
- Relationship & Loss: [Compare, explain, and conclude]. So the answer is yes (or is no).
- Career & Education: [Compare, explain, and conclude]. So the answer is yes (or is no).
- Financial & Legal & Societal: [Compare, explain, and conclude]. So the answer is yes (or is no).
- Lifestyle & Identity & Environment: [Compare, explain, and conclude]. So the answer is yes (or is no).
- None: [Compare, explain, and conclude]. So the answer is yes (or is no).

Answer:
Output exact category names: Mental Health, Physical Health, Abuse & Addiction, Relationship & Loss, Career & Education, Financial & Legal & Societal, Lifestyle & Identity & Environment, or None. Use commas to separate multiple labels, if any.
\end{lstlisting}

\subsection{Well-being Score}
The prompt for the well-being score prediction is shown below, with an example of how debate transcripts are added as enrichment information. The definition of the well-being scale follows \citet{tseriotou-etal-2025-overview}.
\begin{lstlisting}[basicstyle=\footnotesize,breaklines=true,breakatwhitespace=true, frame=single]
Your goal is to analyze and score the following social media post according to the well-being scale below.

Well-being Scale:
- 10: No symptoms and superior functioning in a wide range of activities.
- 9: Absent or minimal symptoms (e.g., mild anxiety before an exam), good functioning in all areas, interested and involved in a wide range of activities.
- 8: If symptoms are present, they are temporary and expected reactions to psychosocial stressors (e.g., difficulty concentrating after a family argument). Slight impairment in social, occupational or school functioning.
- 7: Mild symptoms (e.g., depressed mood and mild insomnia) or some difficulty in social, occupational, or school functioning, but is generally functioning well and has some meaningful interpersonal relationships.
- 6: Moderate symptoms (e.g., panic attacks) or moderate difficulty in social, occupational or school functioning.
- 5: Serious symptoms (e.g., suicidal thoughts, severe compulsions) or serious impairment in social, occupational, or school functioning (e.g., no friends, inability to keep a job).
- 4: Some impairment in reality testing or communication, or major impairment in multiple areas (withdrawal from social ties, inability to work, neglecting family, severe mood/thought impairment).
- 3: A person experiences delusions or hallucinations or serious impairment in communication or judgment or is unable to function in almost all areas (e.g., no job, home, or friends).
- 2: In danger of hurting self or others (e.g., suicide attempts; frequently violent; manic excitement), or may fail to maintain minimal personal hygiene, or has significant impairment in communication (e.g., incoherent or mute).
- 1: The person is in persistent danger of severely hurting self or others, or is persistently unable to maintain minimal personal hygiene, or has attempted a serious suicidal act with a clear expectation of death.

Instructions:
Analyze the post carefully and assign a well-being score. Give an explanation before your final answer.

Post:
"{post}"

Auxiliary information relevant to the post (do not output this):
Debate transcript on possible life event(s):
agent_0:
Round 1:
{response_with_confidence}
Round 2:
{response_with_confidence}

agent_1:
Round 1:
{response_with_confidence}
Round 2:
{response_with_confidence}

Please follow the exact format below.

Output Format:
Explanation:
- Provide step-by-step reasoning to justify the final decision. Do not skip to the answer directly.
Answer:
<score only>
\end{lstlisting}

\subsection{Sharenting Risk}
The prompt for the sharenting risk prediction is shown below, with an example of how predicted labels are added as enrichment information.
\begin{lstlisting}[basicstyle=\footnotesize,breaklines=true,breakatwhitespace=true, frame=single]
You are a Sharenting Risk Classifier. Given a post, classify it into one of four risk levels — A, B, C, or D — based on the degree of information disclosed about a specific child.

Sharenting is when someone shares personal or sensitive information about a specific child (under 18). It also includes posts that might cause harm to the child or to the child's family's reputation. Self-disclosure about oneself, even when under 18, is not considered sharenting.
If the age of the person whose information is being shared is unknown but familial or social roles are mentioned that signal they are a child (e.g., son, daughter, sibling, nephew, niece, grandchild, schoolchild), assume they are children. If they are aged between 18 and 24 (inclusive), sharenting applies only when they are described in a child role within the family or social context.

Sharenting Risk Levels:
{Risk Label Definitions in C.5}

Instructions:
Analyse the post carefully and assign a sharenting risk level (A, B, C, or D). Give an explanation before your final answer.

Post:
"{post}"

Auxiliary information relevant to the post (do not output this):
Possible sharenting behaviour(s) within the following categories: {predicted_label_names}

Classification Logic:
Follow this order strictly: 
1. Check for any A → output "A". 
2. If not, check for any B → output "B". 
3. If not, check for any C → output "C". 
4. Otherwise output "D".
5. Use the highest applicable level if multiple apply.
6. Except that, if the person being disclosed is 18-24 years old and described in a child role, classify directly as C (Low risk), regardless of other disclosures (even if A or B indicators are present).

Please follow the exact format below.

Output Format:
Explanation:
- Provide step-by-step reasoning to justify the final decision. Do not skip to the answer directly.
Answer:
Output only the risk level letter (A, B, C, or D).
\end{lstlisting}

\section{Tables}
Table~\ref{tab:sharenting_failure} breaks down the sharenting risk predictions of the best-performing configuration, CFD transcripts (R), by true risk level, reporting the recall for each level and its most frequent misclassification.
\begin{table}[bh]
\centering
\small
\setlength{\tabcolsep}{4pt}
\begin{tabular}{lccc}
\toprule
True risk & $n$ & Recall (\%) & Most frequent error \\
\midrule
A (high)     & 118 & 58.5 & $\rightarrow$C (33.7) \\
B (moderate) & 62  & 68.8 & $\rightarrow$C (9.3) \\
C (low)      & 84  & 62.7 & $\rightarrow$B (18.0) \\
D (none)     & 117 & 78.1 & $\rightarrow$C (16.0) \\
\bottomrule
\end{tabular}
\caption{Sharenting risk prediction errors by true risk level, using CFD transcripts (R). $n$: number of posts at each risk level. Most frequent error: the most common wrong prediction and its count. Recall and error counts are averaged over 3 seeds.}
\label{tab:sharenting_failure}
\end{table}


\end{document}